\documentclass[british,english,11pt,a4paper]{article}
\usepackage[latin9]{inputenc}
\usepackage{color}
\usepackage{array}
\usepackage{verbatim}
\usepackage{float}
\usepackage{units}
\usepackage{textcomp}
\usepackage{multirow}
\usepackage{amsbsy}
\usepackage{amstext}
\usepackage{amssymb}
\usepackage{graphicx}
\usepackage{esint}

\makeatletter

\newcommand{\noun}[1]{\textsc{#1}}
\providecommand{\tabularnewline}{\\}
\floatstyle{ruled}
\newfloat{algorithm}{tbp}{loa}
\providecommand{\algorithmname}{Algorithm}
\floatname{algorithm}{\protect\algorithmname}

 \usepackage{algolyx}

\usepackage{graphics,color}

\usepackage{colortbl} 
\definecolor{headerColor}{rgb}{0.74,0.88,0.91}
\definecolor{yearPlanColor}{rgb}{0.85,0.93,0.95}

\date{}

\@ifundefined{showcaptionsetup}{}{%
 \PassOptionsToPackage{caption=false}{subfig}}
\usepackage{subfig}
\makeatother

\usepackage{babel}
\addto\captionsbritish{\renewcommand{\algorithmname}{Algorithm}}

\begin{document}

\title{\includegraphics[width=1\columnwidth]{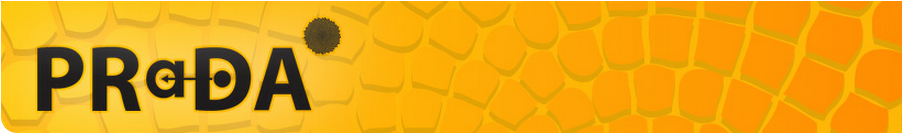}\vfill{}
\textbf{Statistical Latent Space Approach for Mixed Data Modelling
and Applications}\\
\textbf{}\\
}

\author{\textbf{Tu Dinh Nguyen$\dagger$, Truyen Tran$\dagger$$\ddagger$,
Dinh Phung$\dagger$ and Svetha Venkatesh}$\dagger$\\
$\dagger$Center for Pattern Recognition and Data Analytics\\
School of Information Technology, Deakin University, Geelong, Australia\\
$\ddagger$Institute for Multi-Sensor Processing and Content Analysis\\
Curtin University, Australia\\
Email: \{ngtu,truyen.tran,dinh.phung,svetha.venkatesh\}@deakin.edu.au}

\maketitle
\noindent \begin{center}
\vfill{}

\par\end{center}

\begin{tabular}{>{\raggedright}b{0.65\paperwidth}c}
\textsc{\textcolor{blue}{\LARGE{}P}}\textsc{\LARGE{}attern }\textsc{\textcolor{blue}{\LARGE{}R}}\textsc{\LARGE{}ecognition
}\textsc{\textcolor{blue}{\noun{\LARGE{}a}}}\textsc{\LARGE{}nd }\textsc{\textcolor{blue}{\LARGE{}D}}\textsc{\LARGE{}ata
}\textsc{\textcolor{blue}{\LARGE{}A}}\textsc{\LARGE{}nalytics}\\
School of Information Technology, Deakin University, Australia.\\
Locked Bag 20000, Geelong VIC 3220, Australia.\\
Tel: +61 3 5227 2150 \\
Internal report number: \textsf{\textcolor{blue}{\small{}TR-PRaDA-05/13}},
May, 2013. & \includegraphics[scale=0.5]{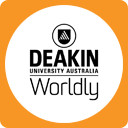}\tabularnewline
\end{tabular}

\thispagestyle{empty}

\global\long\def\realset{\mathbb{R}}

\global\long\def\E{\mathbb{E}}

\global\long\def\realn{\real^{n}}

\global\long\def\natset{\integerset}

\global\long\def\interger{\integerset}

\global\long\def\integerset{\mathbb{Z}}

\global\long\def\natn{\natset^{n}}

\global\long\def\rational{\mathbb{Q}}

\global\long\def\realPlusn{\mathbb{R_{+}^{n}}}

\global\long\def\comp{\complexset}
 \global\long\def\complexset{\mathbb{C}}

\global\long\def\dataset{\mathcal{D}}

\global\long\def\class{\mathcal{C}}

\global\long\def\likelihood{\mathcal{L}}

\global\long\def\normal{\mathcal{N}}

\global\long\def\bphi{\boldsymbol{\phi}}

\global\long\def\bx{\boldsymbol{x}}

\global\long\def\by{\boldsymbol{y}}

\global\long\def\bt{\boldsymbol{t}}

\global\long\def\bf{\boldsymbol{f}}

\global\long\def\bX{\boldsymbol{X}}

\global\long\def\bh{\boldsymbol{h}}

\global\long\def\bv{\boldsymbol{v}}

\global\long\def\ba{\boldsymbol{a}}

\global\long\def\bA{\boldsymbol{A}}

\global\long\def\bb{\boldsymbol{b}}

\global\long\def\bB{\boldsymbol{B}}

\global\long\def\bw{\boldsymbol{w}}

\global\long\def\bW{\boldsymbol{W}}

\global\long\def\bz{\boldsymbol{z}}

\global\long\def\bu{\boldsymbol{u}}

\global\long\def\bI{\boldsymbol{I}}

\global\long\def\bmu{\boldsymbol{\mu}}

\global\long\def\balpha{\boldsymbol{\alpha}}

\global\long\def\bSigma{\boldsymbol{\Sigma}}

\global\long\def\grad{\bigtriangleup}
 
\pagebreak{}

\noindent \begin{center}
\textbf{\LARGE{}Statistical Latent Space Approach for Mixed Data Modelling
and Applications}\\
\textbf{\LARGE{}}\\

\par\end{center}{\LARGE \par}

\noindent \begin{center}
\textbf{Tu Dinh Nguyen$\dagger$, Truyen Tran$\dagger$$\ddagger$,
Dinh Phung$\dagger$ and Svetha Venkatesh}$\dagger$
\par\end{center}

\noindent \begin{center}
$\dagger$Center for Pattern Recognition and Data Analytics
\par\end{center}

\noindent \begin{center}
School of Information Technology, Deakin University, Geelong, Australia
\par\end{center}

\noindent \begin{center}
$\ddagger$Institute for Multi-Sensor Processing and Content Analysis
\par\end{center}

\noindent \begin{center}
Curtin University, Australia
\par\end{center}

\noindent \begin{center}
Email: \{ngtu,truyen.tran,dinh.phung,svetha.venkatesh\}@deakin.edu.au
\par\end{center}

\bigskip{}

\begin{abstract}
The analysis of mixed data has been raising challenges in statistics
and machine learning. One of two most prominent challenges is to develop
new statistical techniques and methodologies to effectively handle
mixed data by making the data less heterogeneous with minimum loss
of information. The other challenge is that such methods must be able
to apply in large-scale tasks when dealing with huge amount of mixed
data. To tackle these challenges, we introduce parameter sharing and
balancing extensions to our recent model, the mixed-variate restricted
Boltzmann machine (MV.RBM) which can transform heterogeneous data
into homogeneous representation. We also integrate structured sparsity
and distance metric learning into RBM-based models. Our proposed methods
are applied in various applications including \emph{latent patient
profile} modelling in medical data analysis and representation learning
for image retrieval. The experimental results demonstrate the models
perform better than baseline methods in medical data and outperform
state-of-the-art rivals in image dataset.
\end{abstract}

\section{Introduction\label{sec:intro}}

\noindent Data can be collected from numerous sources such as cameras,
sensors, human opinions, user preferences, recommendations, ratings,
moods, which together are called \emph{multisource}. Such data can
also exist in any forms of text, hypertext, image, graphics, video,
speech and audio, namely \emph{multimodality}. Data also vary in types
(e.g., binary, categorical, multicategorical, continuous, ordinal,
count, rank) which are known as \emph{multitype}. The combination
of multisource, multimodality and multitype has created \emph{mixed
data} with the characteristics of complexity and heterogeneity. Nowadays
the advances of information technology has enabled seemingly limitless
collection, archive and retrieval of data, leading to an enormous
amount of mixed data. Such huge databases are useful only if they
are analysed to reveal their latent trends, internal dependencies
and structures. Therefore, it is imperative to develop new statistical
techniques and methodologies to effectively handle mixed data. However,
the challenges arise in such tasks are manifolds. One of the two most
prominent challenges is that no simple scaling and transforming methods
can make data less heterogeneous without distorting them. The other
challenge is that the analysis method must be scalable to meet the
requirements of computational complexity and timing when dealing with
huge amount of mixed data. There have been three main research directions
to analyse mixed data as follows:
\begin{itemize}
\item Directly applying machine learning techniques without transforming
the data. In this approach, machine learning methods perform on either
raw data or features extracted from them. Note that these features
are extracted from feature selection frameworks with or without prior
knowledge. Original data and extracted features can be preprocessed
by scaling or normalising, but not projecting to other spaces.
\item Non-probabilistically transforming data to higher-level representations
before performing machine learning methods. Here, researchers use
several methods (e.g., dimensionality reduction techniques) to transform
raw data or features extracted from them into another space. The projected
data on this new space are called higher-level representations of
the original data. These representations are then fed into machine
learning methods for final processing. However, widely-used transformation
methods usually perform in linear fashion without integrating probabilistic
properties.
\item Modelling mixed data using latent variable frameworks. In this case,
probabilistic graphical models are constructed with latent and observed
variables which are obtained from data. These models define joint
distribution over all variables or conditional distribution of latent
variables given observed ones. After that, they can discover latent
representations for the data by computing the conditional probabilities
of latent variables given observed ones. Machine learning methods
then can be applied to these latent representations, similarly to
higher-level representation in the second approach above. Meanwhile,
some latent variable models are highly flexible. They can do classification,
regression, data completion, prediction or even visualisation without
further processing \cite{truyen_phung_venkatesh_fusion12}.
\end{itemize}
\noindent The significant drawback of the first approach is that almost
fundamental methods are only able to handle univariate data or correlation
among a few features at the same time. Some of them can perform on
the combination (e.g., merely concatenated feature vectors) of multiple
features but provide poor results. More interestingly, there are still
several methods which can simultaneously treat multiple types of data
separately and then form a \emph{fusion} at a certain stage. However,
the fusion fails to either flexibly tackle with each type or capture
inter-feature relations. The second approach supports understanding
data distributions, discovering latent structures and latent semantics
of data. Unfortunately, it still depends on the fusion of higher-level
representations when dealing with mixed data. By contrast, models
in the last approach state their outstanding performances. They simultaneously
model multiple types and modalities in probabilistic sense, uncover
latent structures and factors of data, and even efficiently do further
machine learning tasks by themselves. According to these reasons,
we propose our models following the last approach.

The rest of technical report is organised as follows. The next section
explains three main approaches to analyse mixed data by reviewing
typical literature on such approaches. We then describe our approach
and modelling frameworks in Section~\ref{sec:aims_approach}. Next,
Section~\ref{sec:apps} presents the applications of our models on
real datasets. We then discuss about several drawbacks of our model
and build future research plans to overcome those hurdles in Section~\ref{sec:research_proposal}.
Finally, Section~\ref{sec:conclusion} concludes the report.

\section{Mixed Data Analysis: An overview\label{sec:literature_review}}

In this section, we present a literature review on three main approaches
of mixed data analysis. Firstly, we briefly revise several applications
with multitype and multimodality data using the first and second approach,
i.e., using original features and mapping to higher-level representations.
Then we focus on the third approach of using latent variable graphical
models for mixed data modelling.

\subsection{Original features\label{sub:original_feature}}

\noindent Most existing machine learning methods are restricted to
handle univariate data or correlation among a few features. They are
often theoretically designed to deal with single type and modality
of data at the same time. However, there are still several methods
which can simultaneously treat multiple types and modalities of data
separately and then form a \emph{fusion} at a certain stage. Some
of the most considerable include Support Vector Machines (SVMs) and
Neural Networks. Here, for example, we will explain about the SVMs
in more detail.

Support Vector Machines (SVMs) \cite{cortes_etal_ml95_svm,boser_etal_wclt92_svm}
are a widely-used machine learning method for classification, regression
and other tasks. The SVMs have shown their capabilities in pattern
recognition such as handwritten digit recognition, object recognition,
speaker identification, face detection, face recognition and text
categorisation. The method aims to find the optimal hyperplane to
separate relevant and irrelevant samples. The optimisation is done
by maximising the margin between both sets of samples. Note that the
optimisation idea of SVMs lays the foundation for the development
of new approaches including large-margin methods, distance metric
learning and \foreignlanguage{british}{regularisation} techniques
\cite{shawe_etal_campress04_kernel}.

Let $\dataset=\left\{ \left(\bx_{i},y_{i}\right)\right\} $ denote
the training set containing instance-label pairs $\left(\bx_{i},y_{i}\right)$
with $i=1,...,N$ in which $\bx_{i}$ is $n$-dimensional vector,
i.e., $\bx_{i}\in\mathbb{R}^{n}$, and $y_{i}$ has indicator value,
i.e., $y_{i}\in\left\{ 1,-1\right\} $. The SVMs attempt to look for
the solution of following optimisation problem

\begin{eqnarray*}
\min_{\bw,b,\boldsymbol{\xi}} &  & \frac{1}{2}\bw^{\top}\bw+C\sum_{i=1}^{N}\xi_{i}\\
\textnormal{subject to} &  & y_{i}\left(\bw^{\top}\phi\left(\bx_{i}\right)+b\right)\geq1-\xi_{i},\\
 &  & \xi_{i}\geq0.
\end{eqnarray*}

\noindent where $\phi$ is the mapping function, $\boldsymbol{\xi}$
are error terms and $C>0$ is the penalty parameter of the error terms.
The training sample $\bx_{i}$ is mapped into another space. The transformation
here can be non-linear and the projected space maybe higher or even
infinite dimensionality. Therefore, although the separator, i.e.,
separating hyperplane, is linear in high-dimensional space, it possibly
is non-linear in the original one. Denote by $K\left(\bx_{i},\bx_{j}\right)=\phi\left(\bx_{i}\right)^{\top}\phi\left(\bx_{i}\right)$
the kernel function. So far, there have been numerous kernels for
individual problem or category of data. Four basic kernels should
be mentioned as \foreignlanguage{british}{below}
\begin{itemize}
\item Linear kernel: $K\left(\bx_{i},\bx_{j}\right)=\bx_{i}^{\top}\bx_{j}+c$
where $c$ is an optional constant. The linear kernel is the simplest
kernel function basing on the inner product $\left\langle \bx_{i},\bx_{j}\right\rangle $.
\item Polynomial kernel: $K\left(\bx_{i},\bx_{j}\right)=\left(\gamma\bx_{i}^{\top}\bx_{j}+r\right)^{d}$
in which $r$ is the constant, $d$ is polynomial degree, $\gamma>0$
is the slope parameter. This kernel is appropriate to the data which
can be normalised.
\item Sigmoid kernel: $K\left(\bx_{i},\bx_{j}\right)=\tanh\left(\alpha\bx_{i}^{\top}\bx_{j}+c\right)$
with $\alpha$ is the slope parameter\footnote{The slope parameter $\alpha$ usually sets to $\nicefrac{1}{n}$ with
$n$ is the dimensionality of data.} and $c$ is the intercept constant. This kernel is also known as
Hyperbolic Tangent kernel and Multilayer Perceptron (MLP) kernel.
MLP reminisces about the concept of multilayer perceptron neural network.
Indeed, the SVMs with sigmoid kernel act as two-layer perceptron neural
network.
\item Radial basis function (RBF) kernel: $K\left(\bx_{i},\bx_{j}\right)=\exp\left(-\gamma\parallel\bx_{i}-\bx_{j}\parallel^{2}\right)$
with the adjustable parameter $\gamma>0$. The RBF kernel is a generalisation
of Gaussian kernel in which $\gamma=\nicefrac{1}{2\sigma^{2}}$ with
$\sigma$ is the parameter similar to the standard deviation. This
kernel is recommended to be the first choice for classification using
SVMs \cite{chang_etal_tist11_libsvm_guide}. Nonetheless, the parameter
$\gamma$ significantly affects the performance of SVMs. Hence it
should be carefully fine-tuned.
\end{itemize}
In fact, the SVMs with specified kernel function only well suit particular
data type. However, the challenge of understanding mixed data encourages
researchers extend SVMs to tackle multimedia data with multiple types
and modalities. A natural method is to form a fusion which actually
is a combination of the results of several single procedures. Such
fusion approaches, in general, fall into two strategies: early fusion
and late fusion \cite{snoek_etal_acmmm05_fusion}. The SVMs, in particular,
are proposed with the third manner: kernel-based fusion \cite{ayache_etal_ir2007_svmfusion}.
All three fusion schemes of SVMs learn the correlation of types and
modalities at different representation levels using a classifier.

In early fusion, each modality data is first taken into individual
feature engine to extract unimodal features. The extracted unimodal
features are then combined into a single multimodal representation.
An easy way to form a single representation is that the unimodal feature
vectors are concatenated into a fused multimodal vector \cite{snoek_etal_trecvid09_semantic_vse}.
After the combination step, SVMs are applied to fused vectors. It
can be seen that the early fusion supplies all information from original
sources to SVMs. This manner has two-side effects. It benefits independent
learning of regularities for modalities whilst it forces SVMs to use
the same kernel as well as the same parameters for all types of features.
For instance, choosing the $\sigma$ parameter in Gaussian kernel
of SVMs is optimal for the combination vectors, but not necessarily
so for each modality.

The late fusion also begins with feature extractions of unimodal data.
In the next step, however, multiple SVMs perform separately on the
unimodal features. In contrast to early fusion, this type of fusion
aims to capture the regularities of each modality before merging the
outputs to form a uniform semantic representation. This behaviour
creates an advantage for SVMs late fusion procedure. Unfortunately,
the trade-offs are escalating computational complexity and failure
in correlation of mixed features. In late fusion approach, every modality
requires to be trained by separate SVMs framework and after that,
an addition procedure is needed for learning the combination. Besides,
it is hard to capture the correlation among multimodality features
at the combination step.

The early and late fusion approaches are easy to implement. They can
achieve state-of-the-art results, but not always. In cases of imbalanced
influence of the features, such fashions do not provide better results
than sole classification. Two fusions have their own pros and cons.
A good idea can be inventing a hybrid method to balance two approaches.
Being aware of flexible kernel usage of RBM, researchers studied how
to apply kernel methods to SVMs. From that motivation, the third way,
kernel-based fusion, is introduced in \cite{shawe_etal_campress04_kernel}.
Basically, the idea of kernel-based fusion is to integrate features
at the kernel level. After that, it forms a new kernel using a combining
function and the classifier performs on the new kernel. The integration
of multimodal features at kernel level brings many advantages. First,
similarly to the late fusion, it enables the options to select suitable
kernel for each modality (e.g., histogram intersection kernel for
histograms of colors, string kernel or word sequence kernel for documents).
Second, the combination of unimodal kernels to a certain extent still
keeps the correlation among modalities. Finally, the kernel-based
fusion reduces the complexity by excluding the last classified step
comparing to the late fusion.

\subsection{Higher-level representation\label{sub:higher_level_representation}}

Discovering a good representation for both unitype and mixed data
plays critical role in data analysis. An useful representation reveals
latent structure and latent semantics of data. The common approach
is to transform original data or features extracted from data into
another space which contains \emph{higher-level representation}. Then
further computational methods can be easily applied on that representation
of data. Popular transformations often reduce the dimensionality of
the data. Such transformations are also known as dimensionality reduction
techniques with some notable examples like Principal Component Analysis,
Latent Semantic Indexing, Linear Discriminant Analysis and Non-negative
Matrix Factorisation. 

Principal Component Analysis (PCA) \cite{jolliffe_sprg86_pca} is
probably the most popular technique employed for numerous applications
including dimensionality reduction, lossy data compression, feature
extraction and data visualisation. The PCA learns a linear transformation
which is exactly an orthogonal projection of the data onto a lower
dimensional linear space called ``principal subspace''. The projection
function aims to maximise the variance of projected data on this subspace.
The PCA can discover latent structure of data by examining the projected
data in subspace. Besides, the projection of PCA preserves most information
of the data. Hence it is possible to use the PCA as data compression
tool. This method has been widely used in a number of machine learning
tasks such as face recognition \cite{zhao_etal_sprg98_discriminant},
image compression, object recognition \cite{johnson_etal_pami99_3dscene},
document classification, document retrieval \cite{torkkola_icdm01_linear}
and video classification \cite{xu_etal_icme03_video}.

Deerwester~\emph{et~al}. \cite{deerwester_etal_JASIS90} propose
Latent Semantic Indexing (LSI) to find a linear transformation which
projects a document consisting of word counts onto a latent eigen-space.
This latent space captures the semantics of document. The LSI is originally
invented to handle text data. Another technique is Linear Discriminant
Analysis (LDA) related to Fisher's linear discriminant \cite{fisher_eug36_lda}.
The LDA is assumed to be introduced by Fisher. The technique acts
as a two-class classifier finding a linear combination of features
which best separates the data.

Lastly, Non-negative Matrix Factorisation (NMF) \cite{lee_etal_nature99_nmf}
finds linear representation of non-negative data. It approximately
factorises the data into two non-negative factors which are matrices.
Recall that the aforementioned PCA can be seen as an matrix factorisation
method. But unlike PCA, the NMF obligates both two matrices to be
non-negative. This constraint has several advantages: (i) well-suited
for many applications (e.g., collaborative filtering, count data modelling)
which require the data to be non-negative; (ii) useful for parts-based
representation learning \cite{lee_etal_nature99_nmf}; (iii) providing
purely additive representations without any subtractions.

As can be seen, these dimensionality reduction tools above can be
implemented into various data including image, text, video and count.
When dealing with mixed data, such methods are probably employed for
each modality. After that, the projected features of individual modality
can be fused into a single representation. The concatenation, for
example, is one of the simplest ways to integrate all features. The
concatenated vector now presents the higher-level representation,
completing a step equivalent to the early fusion in Section~\ref{sub:original_feature}.

\subsection{Latent Representation}

Transforming to higher-level representation in Section~\ref{sub:higher_level_representation}
maintains a non-probabilistic mapping whilst the data often follow
a set of probabilistic distributions. That leads to the limitations
of fitting data and scalable capabilities of such methods. Another
approach is to model the data using probabilistic graphical models.
Such models often consist of observed variables representing data
and latent variables which coherently reflect the regularities, structures
and correlations of data. The probabilistic functions of observed
and latent variables map data onto latent space forming \emph{latent
representation} of the data. The latent representation is like higher-level
representation, but obtained with probabilistic mapping manner. As
for handling mixed data, there are several methods such as Bayesian
latent variables models and Restricted Boltzmann Machine, which simultaneously
model multiple types of data, yet others separately treat each type
and then combine representations in fusion fashion. The former methods
prove more powerful than the latter ones. Among these approaches,
Restricted Boltzmann Machine demonstrates its outstanding capabilities
and flexibilities. It is easy to enhance the latent representation
with sparsity and metric learning integration. Moreover, the inference
in this model is much more scalable: each Monte Carlo Markov Chain
sweeping through all variables takes only linear time. It is more
practical to build a model with hundreds of latent variables to deal
with big datasets. In this part, we go through methods using fusion
style and focus on simultaneously modelling methods, especially on
Restricted Boltzmann Machine.

\subsubsection{Separate modelling}

Similarly to mapping into higher-level representation, in this approach,
the models capture data types separately, retrieve latent representations
with projections and then fuse them into combined one. The difference
is that their approaches are probabilistic instead of non-probabilistic.
The methods which are used for producing higher-level representations
are upgraded to probabilistic versions including probabilistic Principal
Component Analysis, probabilistic Latent Semantic Indexing and Bayesian
Nonparametric Factor Analysis. In addition, topic modelling methods
like Latent Dirichlet Allocation has also been widely-used.

Blei~\emph{et~al}. raise the research direction of ``topic modelling''.
Here, a topic is constructed by a group of words which frequently
appear together. In fact, they propose a graphical model which includes
latent variables indicating latent topics and observed ones which
denote words in document. Particularly, they introduce Latent Dirichlet
Allocation (LDA) \cite{blei_etal_jmlr03_lda} to analyse a huge volume
of unlabelled text. Meanwhile, probabilistic Principal Component Analysis
(pPCA) \cite{tipping_etal_jrss99_pPCA} borrows formulation of PCA
to a Gaussian latent variable model. The pPCA reduces the dimensionality
of observed data to a lower dimensionality of latent variable via
a linear transformation function. Another method is probabilistic
Latent Semantic Indexing (pLSI) \cite{hofmann_uai99_pLSI}, an extension
of LSI. It builds a probabilistic framework in which the joint distribution
of document index and sampled words variables are modelled using conditionally
independent multinomial distribution given words' topics. Lastly,
Bayesian Nonparametric Factor Analysis (BNFA) \cite{paisley_etal_icml09_bnfa}
takes recent advances in Bayesian nonparametric factor analysis for
factor decomposition. Comparing to NMF, it overcomes the limitation
of determining the number of latent factors in advance. The BNFA can
automatically identify the number of latent factors from the data.

\subsubsection{Mixed data modelling}

Mixed data has been investigated under various names such as mixed
outcomes, mixed responses, multitype and multimodal data. They have
been studied in many works in statistics \cite{sammel_etal_jrss97_latent,dunson_jrss00_bayesian,shi_etal_jrss00_latent,mcculloch_smmr08_joint,catalano_statmed97_bivariate,song_etal_biomet09_joint,deleon_statmed11_copula,dunson_herring_biostat05_bayesian,murray_amstats13_bayesian}.
In \cite{sammel_etal_jrss97_latent}, they generalise a class of latent
variable models that combines covariates of multiple outcomes using
generalised linear model. The model does not require the outcomes
coming from the same family. It can handle multiple latent variables
that are discrete, categorical or continuous. Similarly, Dunson~\emph{et~al.}
apply separate generalised linear models for each observed variables
\cite{dunson_jrss00_bayesian,dunson_amstats03_dynamic}. Ordinal and
continuous outcomes can also be jointly modelled in another way \cite{catalano_statmed97_bivariate}.
Ordinal data, first, is observed by continuous latent variable. Then,
continuous data and these latent variables offer a joint normal distribution.
Confirmatory factor analysis using Monte Carlo expectation maximisation
is applied to model mixed continuous and polytomous data \cite{shi_etal_jrss00_latent}. 

Generally speaking, the existing literature offers three approaches:
The first is to specify the direct type-specific conditional relationship
between two variables (e.g., see \cite{mcculloch_smmr08_joint}),
the second is to assume that each observable is generated from a latent
variable (latent variables then encode the dependencies) (e.g., see
\cite{dunson_herring_biostat05_bayesian})\textcolor{black}{, and
the third is to construct joint cumulative distributions using copula
\cite{song_etal_biomet09_joint,deleon_statmed11_copula}}. The drawback
of the first approach is that it requires far more domain knowledge
and statistical expertise to design a correct model even for a small
number of variables. The second approach lifts the direct dependencies
to the latent variables level. The inference is much more complicated
because we must sum over all latent variables. It takes exponential
complexity. All approaches are, however, not very scalable to realistic
setting of large-scale datasets. Recently, models using Restricted
Boltzmann Machine offer the fourth alternative: direct pairwise dependencies
are substituted by indirect long-range dependencies. RBM has successfully
captured single type of data and then been developed to handle multiple
data types. It has been shown that the RBM still keeps its powers
when dealing with mixed data. Here we present about the RBM's historical
development.

\paragraph{Restricted Boltzmann Machine.}

Recently, Restricted Boltzmann Machine (RBM) \cite{smolensky_mit86_parallel,freund_haussler_94_unsupervised,hinton_neucom02_training}
has attracted massive attention due to its versatility in unsupervised
and supervised learning tasks. The model has been spreading all over
numerous research fields. The RBM can model multivariate distribution,
perform feature extraction, classification, data completion, prediction
and construct deep architectures \cite{hinton_salakhutdinov_sci06_reducing}.
In model representation, RBM is a bipartite undirected graphical model
with two layers. The first layer includes observed variables called
visible units while the second layer consists of latent variables,
known as hidden units. The bipartite graph contains only inter-layer
connections without any intra-layer connections. It also means that
there are only pairwise interactions for units between two layers.
The hidden layer describes the latent representation of the data.
Exploiting the joint and conditional distributions between hidden
and visible layers can discover latent structure of given data. Thanks
to the restricted connections, the hidden posteriors and the probability
of the data are easy to evaluate. It is possible to extract features
quickly and do sampling-based inference efficiently \cite{hinton_neucom02_training}.

The RBM was developed strongly at the dawn of its birth. Nevertheless,
most works of RBM at that time were applied to unitype and unimodality
which are individual type of data. The most popularly used were \emph{binary}
\cite{freund_haussler_94_unsupervised} and \emph{Gaussian} \cite{hinton_salakhutdinov_sci06_reducing}.
In \cite{freund_haussler_94_unsupervised}, they simplified the Boltzmann
Machine \textcolor{red}{\cite{ackley_etal_cogsci85_bm}} by restricting
the graph to a bipartite form, creating the so-called influence combination
machine or the combination machine. Their model is closely similar
to the Harmonium model studied by Smolensky~\emph{et~al}. \cite{smolensky_mit86_parallel}.
They applied the model to synthesised binary image and NIST handwritten
dataset\footnote{NIST is the larger set of MNIST dataset \cite{lecun_etal_ieee98gradient}}.
\emph{Binomial} units are introduced in \cite{teh_etal_nips01} as
an improvement of binary units. Each binomial unit comprises more
information than ordinary binary unit. In fact, the new unit is a
finite series of copies of the old ones. These copies share the same
set of weights and biases. It is an important property because the
weight-sharing setting leaves the mathematics unchanged. Thus, model
is adjusted but probabilistic equations and learning remain unchanged.
In \cite{nair_hinton_icml10_rectified}, they generalised RBM with
binary stochastic hidden units by infinite number of copies that share
the same weights. Similarly to the binomial unit in \cite{teh_etal_nips01},
the series of these units has no more parameters than an ordinary
binary unit. The difference of this series from binomial series is
that it is infinite. Hence, it exhibits much more information. This
type of hidden units is called ``\emph{rectified}'' units. Because
the rectified linear units have zero biases and are noise-free, they
can capture intensity equivariant. The output of RBM with rectified
units are invariant. They do not depend on the variants of input such
as scale, orientation and light condition.

The RBM with Gaussian type was also popular at the beginning period.
It was first applied to pixel intensities of images \cite{hinton_salakhutdinov_sci06_reducing}.
A deep autoencoder was constructed to encode and decode handwritten
digits in MNIST dataset \cite{lecun_etal_ieee98gradient}. The input
data acted as linear input units with independent Gaussian noise.
Basing on Gaussian RBM, another work proposed mean-covariance RBM
abbreviated as mcRBM \cite{ranzato_hinton_cvpr10_mcRBM}. This model
combined two approaches: probabilistic and hierarchical models to
learn representation of images. It could learn features on natural
image patches and use them to recognise object categories, outperforming
state-of-the-art methods by its date. Later, Gaussian RBM-based model
became able to handle vector-variate and matrix-variate ordinal data
\cite{truyen_phung_venkatesh_acml12a}. The improved model, Cumulative
RBM is different from its precedent because its utilities are never
fully observed. The model obtained competitive results against state-of-the-art
rivals in collaborative filtering using large-scale public datasets.
The work's contributions are, firstly, to enhance RBM model, and secondly,
to extend the application on multivariate ordinal data from modelling
i.i.d. vectors to modelling matrices of correlated entries.

Salakhutdinov~\emph{et~al}. \cite{salakhutdinov_etal_icml07_rbmcf}
consider the data collected from user ratings of movies and aim to
solve collaborative filtering problem. They successfully model tabular
data such as user ratings of movies using RBM. The empirical performance
of their model gains slightly better result than carefully-tuned Singular
Value Decomposition (SVD) models on Netflix dataset. Moreover, when
they linearly combine multiple RBMs and SVD models, they achieve an
error rate 6\% lower than the score of Netflix's own system. In their
model, they treat each user rate for a movie as categorical data.
More specifically, they use a conditional multinomial (a ``softmax'')
distribution to model each column which is a binary vector with only
one elements equal to 1 and the rest equal to 0. This is the first
attempt of using RBM to model \emph{categorical} data. Also solving
collaborative filtering problem, Truyen~\emph{et~al}. pay more attention
to item-based and user-based correlations. This behaviour is different
from the work in \cite{salakhutdinov_etal_icml07_rbmcf} in which
RBM is constructed for each user and his ratings. The correlations
lead to a general mode of Boltzmann Machine and the model must handle
the ordinal nature of ratings. Thus, they propose \emph{ordinal} Boltzmann
Machine for joint modelling user-based and item-based processes.

The \emph{Poisson} distribution is employed into RBM model to deal
with text data \cite{gehler_etal_icml06}. Documents are considered
as bags of words which represent count data. The count data is observed
by visible layer of RBM using conditional Poisson distributions. The
hidden layer consists of latent topic variables which are modelled
by conditional binomial distributions. This model, called rate adapting
Poisson (RAP), can be considered as a generalised version of exponential
family harmonium model \cite{welling_etal_nips05_exponential} which
uses multinomial conditional distributions to model the input data.
However, the RAP assumes Poisson rates for all documents are the same.
It can not deal with different lengths of document. Overcoming that
limitation, Constrained Poisson RBM is presented in \cite{salakhutdinov_hinton_SIGIR07_semantic}.
In this work, they introduce a constraint to Poisson RBM to guarantee
that the sum of mean Poisson rates over entire words is always equal
to the length of the document. Though more flexible, its trade-off
is that it can not define a proper probability distribution over the
word counts. They, again, propose to totally share parameters, i.e.,
visible biases, hidden biases and inter-connections weights, of repeated
words \cite{salakhutdinov_hinton_nips09_repsoftmax}. The visible
units are now modelled by softmax distributions in the so-called ``replicated
softmax'' model. It provides a more stable and better way of tackling
different lengths of documents.

Le Roux~\emph{et~al}. propose an improved model of Gaussian RBM,
called betaRBM, which defines \emph{beta} distributions on conditional
distributions of visible variables given hidden ones \cite{leroux_etal_neucom11learning}.
In this model, the variances of Gaussian noises are not fixed. The
betaRBM learns both means and variances during training phase. Unfortunately,
this makes training of betaRBM intractable. Therefore, they slightly
adjust the energy function to weaken the constraints of parameters
of hidden units without losing proper distributions.

As for other modalities such as voice, audio, motion and video, there
are also many works exploiting RBM. The most remarkable are two series
of Taylor~\emph{et~al}. on video, i.e., human motion recognition
\cite{taylor_etal_nips06,taylor_etal_jmlr11_time} and Mohamed~\emph{et~al}.
on audio, i.e., speech recognition \cite{mohamed_etal_2012_acoustic_dbn,mohamed_etal_icassp10,mohamed_etal_nips09_dbn_pr}.
There are temporal dependencies inside video and audio, like sequences
of frames from time to time in video. Each static frame in video are
modelled by Gaussian RBM. To deal with modelling temporal information,
the so-called \emph{conditional} RBM (CRBM) are employed to take visible
variables of previous time slides as additional input for current
time slide \cite{taylor_etal_nips06}. These previous slides connect
to hidden layers and current visible configuration using directed
connections. The additional connecting does not make inference more
difficult. Mohamed~\emph{et~al.} apply CRBM to speech recognition,
in which they represent the speech using sequence of Mel-Cepstral
coefficients.

In \cite{xing_etal_uai05_dualwing}, Dual-Wing RBM is introduced to
model simultaneously continuous and Poisson variables. In this approach,
the hidden units are similar to ``latent topics'' of Latent Dirichlet
Allocation (LDA) \cite{blei_etal_jmlr03_lda}. However, the connections
of Dual-Wing RBM are undirected whilst those of LDA are directed.
Therefore, it can model joint distribution instead of conditional
distribution given the input data. Later, the Mixed-Variate RBM (MV.RBM)
is the first to enclose six data types in a single model \cite{truyen_phung_venkatesh_jmlr11_mvrbm}.
In fact, MV.RBM can be viewed as a generalisation of RBM for modelling
multiple types and modalities. MV.RBM still keeps all capabilities
of performing a number of machine learning tasks including feature
extraction, dimensionality reduction, data completion and prediction.

\subsubsection{Structured sparsity and Metric Learning}

With classification task, there should be only small relevant parts
of latent factors of the data. This characteristic calls for the sparsity
in latent representation. In sparse latent representation, the features
are not intensive and often equal to zero \cite{olshausen_etal_nature96_emergence,olshausen_etal_vision97_sparse}.
A regularisation is a common choice to control the sparsity of latent
representation. For example, Lee~\emph{et~al}. add a regularisation
term to force the deviation of expected activation of hidden units
in RBM to a specified level \cite{lee_etal_nips08_sparse}. This allows
only a small number of hidden units to be activated. The $\nicefrac{l_{1}}{l_{2}}$
regulariser known as group lasso has attracted much interest from
both statistics community and machine learning community. However,
this regulariser can not control the intra-group sparsity. More recently,
Luo~\emph{et~al}. introduce a mixed-norm $\nicefrac{l_{1}}{l_{2}}$
regulariser which can control both inter-group and intra-group sparsities
\cite{luo_etal_aaai11_sgrbm}. This is a combination of $l_{1}$ and
$l_{2}$ normaliser which are integrated into Boltzmann machines.
The integration shows the efficiency of the regulariser on classification
task of handwritten digits.

Metric learning provides better representations for subsequent machine
learning tasks such as classification and retrieval. There are some
works which attempt to solve distance function problem \cite{hertz_etal_cvpr04_distfunc,goldberger_etal_nips04_nca,yu_etal_icme06_adaboosted,tao_etal_ieee06_lda,weinberger_etal_nips06_lmnnc,truyen_phung_venkatesh_icme12_mlrbm}.
They follow two common approaches: (i) creating a new distance function
from classifiers; (ii) improving distance function for retrieval technique
such as $k$-NN. For example, Goldberger~\emph{et~al}. \cite{goldberger_etal_nips04_nca}
propose Neighbourhood Component Analysis (NCA) which is a distance
metric learning algorithm to improve $k$-NN classification. Mahalanobis
distance is improved by minimising the leave-one-out cross validation
error. Unfortunately, it is difficult to compute the gradient because
it uses softmax activation function to incorporate the probabilistic
property into distance function. Weinberger~\emph{et~al}. \cite{weinberger_etal_nips06_lmnnc}
introduce the so-called ``large margin nearest neighbour'' in which
the objective function consists of two terms. The first term is to
shorten the distance between each input and its target neighbours
whilst the second term is to enlarge the gap from each point to different
labeled ones. This method thereby requires labelled samples. The idea
of shortening intra-concept distances is implemented in learning phase
of RBM \cite{truyen_phung_venkatesh_icme12_mlrbm}. An information-theoretic
distance metric known as symmetrized Kullback-Leibler divergence is
used as a regulariser of objective function during training time of
RBM. This method supplies new representations which can capture the
regularities and invariance of human faces.

\section{Latent Space Approaches for Mixed Data Modelling\label{sec:aims_approach}}

Throughout Section~\ref{sec:literature_review}, the graphical model
with latent variables is the most promising method to discover latent
representation of mixed data. This research also aims at parts of
ongoing effort to model multivariate data. Our main approaches are
to apply Restricted Boltzmann Machine (RBM) and its variants, explore
structured sparsity and learn an information-theoretic distance on
latent representations which are mapped by RBMs. In this section,
we summarise the background of RBM, Mixed-Variate RBM (MV.RBM) and
how to integrate structured sparsity and distance metric learning
into such frameworks.

\subsection{Restricted Boltzmann Machine\label{sub:RBM}}

A Restricted Boltzmann Machine (RBM) is a bipartite undirected graphical
model with two layers \cite{smolensky_mit86_parallel,freund_haussler_94_unsupervised,hinton_neucom02_training}.
The first layer contains observed variables as called as visible units
while the second layer consists of latent variables, known as hidden
units. Inheriting from Boltzmann Machine \cite{ackley_etal_cogsci85_bm},
RBM is a particular form of log-linear Markov Random Field (MRF) \cite{lauritzen_oxford96_graphical,darroch_etal_astat80_markov}.
The difference between RBM and general Boltzmann Machine is that RBM
only contains inter-layer connections among visible and hidden units.
It is as equivalent to that intra-layer connections which are connections
among visible-visible, hidden-hidden units are excluded. The Figure~\ref{fig:RBM_graphical_model}
demonstrates an example of RBM with 4 visible units and 3 hidden units
in form of graphical model.

\noindent \begin{center}
\begin{figure}[h]
\noindent \begin{centering}
\includegraphics[scale=0.15]{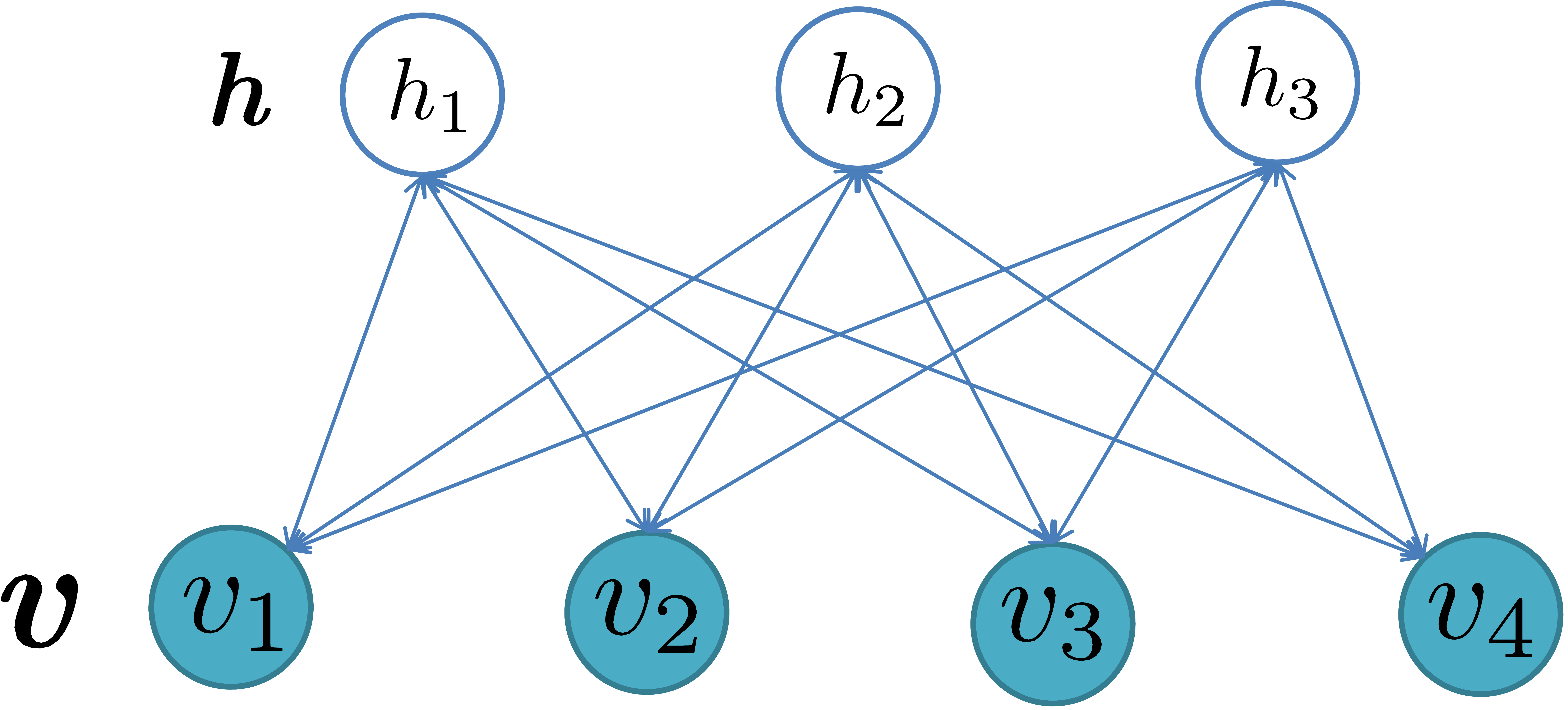}
\par\end{centering}

\centering{}\caption{Graphical model of a RBM that models joint distribution of 4 visible
units and 3 hidden units. The connections are undirected and the shade
nodes are observed.\label{fig:RBM_graphical_model}}
\end{figure}

\par\end{center}

\subsubsection{Model representation}

Let $\bv$ denote the set of visible variables: $\bv=\left\{ v_{1},v_{2},...,v_{N}\right\} $,
$\bh$ indicate the set of hidden ones: $\bh=\left\{ h_{1},h_{2},...,h_{K}\right\} $.
RBM assigns probabilities to variables basing on energy theory \cite{hopfield_nasci82_neural}.
Thus, RBM is a energy-based model which captures dependencies between
variables by attaching a scalar energy, a measure of compatibility,
to each \emph{configuration}\textbf{ }of the variables. The learning
phase of energy-based model seeks an energy function that minimises
energies of correct values and maximises for incorrect values \cite{lecun_etal_psd06_tutorial}.
In RBM case, energy function of a joint configuration $\left(\bv,\bh\right)$
is given by

\[
E\left(\bv,\bh;\psi\right)=-\boldsymbol{a}^{\top}\bv-\boldsymbol{b}^{\top}\bh-\bh^{\top}\boldsymbol{W}\bv
\]

where $\psi=\left\{ \boldsymbol{a}\mathbf{,}\boldsymbol{b},\boldsymbol{W}\right\} $
is the set of parameters with $\boldsymbol{a},\boldsymbol{b}$ are
biases of hidden and visible units, respectively, and $\boldsymbol{W}$
represents the weights connecting hidden and visible units. The network
assigns a probability to every pairwise configuration

\begin{equation}
P\left(\bv,\bh;\psi\right)=\frac{1}{\mathcal{Z}\left(\psi\right)}e^{-E\left(\bv,\bh;\psi\right)}\label{eq:RBM_Prob}
\end{equation}

Note that $\mathcal{Z}\left(\psi\right)$ express the \emph{partition
function} or \emph{normalise factor}, given by summing over all possible
configuration pairs of visible and hidden units

\begin{equation}
\mathcal{Z}\left(\psi\right)=\sum_{\bv,\bh}e^{-E\left(\bv,\bh;\psi\right)}\label{eq:RBM_Part}
\end{equation}

Marginalizing out $\bh$, we have the probability that model assigns
to visible configuration $\bv$

\[
P\left(\bv;\psi\right)=\sum_{\bh}P\left(\bv,\bh;\psi\right)=\frac{1}{\mathcal{Z}\left(\psi\right)}\sum_{\bh}e^{-E\left(\bv,\bh;\psi\right)}
\]

Interestingly, RBM itself possesses a striking property of dependence
among variables. Since RBM has no intra-layer connections, every pairs
of units in each layer suffer conditional independence on another
layer. Hence, the conditional distributions over hidden and visible
units are factorised as

\begin{eqnarray}
P\left(\bv\mid\bh;\psi\right) & = & \prod_{i=1}^{N}P\left(v_{i}\mid\bh\right)\label{eq:RBM_factorVis}\\
P\left(\bh\mid\bv;\psi\right) & = & \prod_{j=1}^{K}P\left(h_{j}\mid\bv\right)\label{eq:RBM_factorHid}
\end{eqnarray}

\subsubsection{Parameter Learning}

As mentioned above, the model adjusts parameters $\psi=\left\{ \ba,\bb,\bW\right\} $
to minimise energies as equivalent to maximise the probabilities assigned
to data. The data likelihood is marginal distribution of visible configuration
\textbf{$\bv$}: $\mathcal{L}=P\left(\bv;\psi\right)=\sum_{\bh}P\left(\bv,\bh;\psi\right)$.
Parameter estimation might be performed by using \emph{gradient ascent}
to maximise likelihood. Apparently, another description of Eq.~(\ref{eq:RBM_Prob})
is as below 
\begin{eqnarray}
P\left(\bv,\bh;\psi\right) & = & \exp\left\{ -E\left(\bv,\bh;\psi\right)-\log\mathcal{Z}\left(\mathbf{\psi}\right)\right\} \nonumber \\
 & = & \exp\left\{ \ba^{\top}\bv+\bb^{\top}\bh+\bh^{\top}\bW\bv-A\left(\psi\right)\right\} \label{eq:RBM_ExpProb}
\end{eqnarray}
where $A\left(\psi\right)=\log\mathcal{Z}\left(\psi\right)$ is log-partition
function. It is clear that Eq.~(\ref{eq:RBM_ExpProb}) forms an exponential
function. Therefore, RBM belongs to exponential family. The gradient
of $\log\mathcal{L}$ with respect to parameters holds the type of
difference of expectations

\begin{equation}
\frac{\partial}{\partial\psi}\left(\log\mathcal{L}\left(\bv;\psi\right)\right)=\E_{\bv,\bh}\left[\frac{\partial E\left(\bv,\bh\right)}{\partial\psi}\right]-\E_{\bh\mid\bv}\left[\frac{\partial E\left(\bv,\bh\right)}{\partial\psi}\right]\label{eq:RBM_logLLGrad}
\end{equation}

$\E_{\bv,\bh}$ describes the expectation with respect to full model
distribution $P\left(\bv,\bh;\psi\right)=P\left(\bh\mid\bv;\psi\right)P\left(\bv;\psi\right)$.
$\E_{\bh\mid\bv}$ denotes the expectation with respect to conditional
distribution given known $\bv$. To be more specific, the derivative
of the log-likelihood with respect to model parameters can be computed
as below
\begin{eqnarray}
\frac{\partial\log\mathcal{L}\left(\bv;\psi\right)}{\partial\bW} & = & \E_{P_{\textnormal{data}}}\left[\bh^{\top}\bv\right]-\E_{P_{\textnormal{model}}}\left[\bh^{\top}\bv\right]\label{eq:paramsWGrad}\\
\frac{\partial\log\mathcal{L}\left(\bv;\psi\right)}{\partial\ba} & = & \E_{P_{\textnormal{data}}}\left[\bv\right]-\E_{P_{\textnormal{model}}}\left[\bv\right]\label{eq:paramsAGrad}\\
\frac{\partial\log\mathcal{L}\left(\bv;\psi\right)}{\partial\bb} & = & \E_{P_{\textnormal{data}}}\left[\bh\right]-\E_{P_{\textnormal{model}}}\left[\bh\right]\label{eq:paramsBGrad}
\end{eqnarray}

Thanks to the factorisation in Eq.~(\ref{eq:RBM_factorHid}), it
is simple to compute the data expectation. However, the model expectation
is intractable to estimate exactly. Consequently, several approximate
methods must be considered. In this report, we choose Markov Chain
Monte Carlo (MCMC) method, one of the most powerful samplers. Due
to factorisations in Eq.~(\ref{eq:RBM_factorVis},\ref{eq:RBM_factorHid}),
MCMC enables us to do sampling and approximate efficiently. More particularly,
alternating Gibbs sampling between hidden and visible units, $\hat{\bv}\sim P\left(\bv\mid\hat{\bh}\right)$
and $\hat{\bh}\sim P\left(\bh\mid\hat{\bv}\right)$, supplies samples
of the equilibrium distribution. On one hand, all hidden and visible
units are updated in parallel in individual iteration of Gibbs sampling.
After estimating states for hidden units, the process changes into
a \emph{reconstruction} step in which visible states are computed
and then sampled on them. The difference between first step and last
reconstruction step shows how much parameters should be changed in
order to conduct data distribution closely to equilibrium distribution.
On the other hand, the learning is able to be pushed to higher speed
by applying Contrastive Divergence (CD) \cite{hinton_neucom02_training},
a truncated MCMC-based method, which in fact is Kullback-Liebler divergence
of two distributions. In Contrastive Divergence, the Markov Chain
is restarted to each observed sample at each stage and then all parameters
are updated. $\textnormal{CD}{}_{n}$ denotes using $n$ full steps
of alternating Gibbs sampling. In practical experiment, $\textnormal{CD}{}_{1}$
is used more frequently since it is proved that Kullback-Liebler divergence
reduces after each step \cite{hinton_neucom02_training}. Especially
with large training dataset, $\textnormal{CD}{}_{1}$ is much more
faster than $\textnormal{CD}{}_{n}$. It also has low variance, but
still provides dramatic difference from equilibrium distribution when
the mixing rate is low. An algorithm called Persistent Contrastive
Divergence (PCD) was introduced in \cite{tieleman_icml08_pcd}. Instead
of resetting Markov Chain between parameters updates, in PCD, they
initialise the chain at the state in which it ended at the previous
one. This way of initialisation contributes fairly closer to model
distribution, even though the model slightly changes after parameters
updates.

Once samples have been supplied, the parameters are updated in a gradient
ascent fashion as follows

\begin{equation}
\psi\leftarrow\psi+\lambda\left(\E_{\bv,\bh}\left[\frac{\partial E\left(\bv,\bh\right)}{\partial\psi}\right]-\E_{\bh\mid\bv}\left[\frac{\partial E\left(\bv,\bh\right)}{\partial\psi}\right]\right)\label{eq:RBM_paramUpdate}
\end{equation}

\noindent for some learning rate $\lambda>0$. Because the expectations
in Eq.~(\ref{eq:RBM_paramUpdate}) can be estimated after taking
each sample or mini-batch of samples, it is more efficient to use
stochastic gradient ascent instead of normal gradient ascent. The
algorithm to learn parameters for RBM using $\textnormal{CD}_{1}$
is described as in Alg.~\ref{alg:RBM_CD}.

\noindent \begin{center}
\begin{algorithm}
\textbf{Input}: 

Sample $\bv_{0}$, learning rate $\lambda$ as in Eq.~(\ref{eq:RBM_paramUpdate}).

$\left\{ \ba,\bb,\bW\right\} $ is the set of parameters of RBM as
in Eq.~(\ref{eq:RBM_ExpProb}).

\rule[0.5ex]{1\columnwidth}{0.05pt}\vspace{-0.005\textheight}

\begin{algor}[1]
\item [{for}] $j=1$ to $K$
\item [{{*}}] $P\left(h_{0j}=1\mid\bv_{0}\right)=\textnormal{sigm}\left(b_{j}+\sum_{i}v_{0i}W_{ij}\right)$
\item [{{*}}] $h_{0j}\sim P\left(h_{0j}=1\mid\bv_{0}\right)$
\item [{endfor}]~
\item [{for}] $i=1$ to $N$
\item [{{*}}] $P\left(v_{1i}=1\mid\bh_{0}\right)=\textnormal{sigm}\left(a_{i}+\sum_{j}h_{0j}W_{ij}\right)$
\item [{{*}}] $v_{1i}\sim P\left(v_{1i}=1\mid\bh_{0}\right)$
\item [{endfor}]~
\item [{for}] $j=1$ to $K$
\item [{{*}}] $P\left(h_{1j}=1\mid\bv_{1}\right)=\textnormal{sigm}\left(b_{j}+\sum_{i}v_{1i}W_{ij}\right)$
\item [{endfor}]~
\item [{{*}}] $\bW\leftarrow\bW+\lambda\left(\bh_{0}\bv_{0}^{\top}-P\left(h_{1j}=1\mid\bv_{1}\right)\bv_{1}^{\top}\right)$
\item [{{*}}] $\ba\leftarrow\ba+\lambda\left(\bv_{0}-\bv_{1}\right)$
\item [{{*}}] $\bb\leftarrow\bb+\lambda\left(\bh_{0}-P\left(h_{1j}=1\mid\bv_{1}\right)\right)$
\end{algor}
\rule[0.5ex]{1\columnwidth}{0.05pt}

\textbf{Output}: $\left\{ \ba,\bb,\bW\right\} $.

\caption{Learning parameters of RBM using Contrastive Divergence.\label{alg:RBM_CD}}
\end{algorithm}

\par\end{center}

Parameter learning of RBM benefits enormously in \emph{deep learning}.
In this field, RBM is considered as an infinite belief networks with
tied weights \cite{hinton_neucom06_fastDBN}. A deep belief network
(DBN) is a hybrid network with multiple layers. Top two layers contain
undirected connections forming an associative memory while the layers
below have directed ones. This kind of deep networks might be pretrained
layer-by-layer with individual two layers forming a RBM. After initialising
weights by using multiple RBMs, all parameters of this hierarchical
model are fine-tuned by back-propagation similar to neural networks.
The advantages of unsupervised pretraining with respect to deep architectures
are proved empirically in \cite{erhan_etal_jmlr10_pretrain}.

\subsubsection{Inference}

The aim of learning RBM is to study a generative model that generates
visible data, also known as training data. This benefits in numerous
applications including dimensionality reduction, feature extraction
\cite{hinton_salakhutdinov_sci06_reducing} and data completion as
well as collaborative filtering \cite{salakhutdinov_etal_icml07_rbmcf}.
For the former one, the posterior projection of RBM is useful for
feature extraction tool and used as new domain of features with lower
dimension. Once the model has been estimated, visible input data can
be transformed into real-valued hidden vector $\hat{\bh}=\left(\hat{h}_{1},\hat{h}_{2},...,\hat{h}_{K}\right)$,
where $\hat{h}_{k}=P\left(h_{k}=1\mid\bv\right)$. It is conspicuous
that handling vector in form of $\hat{\bh}$ are much more convenient
than original data vector $\bv$, especially when the data types are
mixed \cite{truyen_phung_venkatesh_jmlr11_mvrbm} (see Section~\ref{sub:MV.RBM}).
RBM learns data in generative way with whole information in numerous
types condensed in numerical hidden posterior vectors. Consequently,
these numerical vectors are fed into next data analysis tools to do
clustering, classification, prediction, even continue reduce dimension
like multilayer encoder network \cite{hinton_salakhutdinov_sci06_reducing}.
To present evidence for reserving data information, we reconstruct
the original data using $\hat{\bv}=\left(\hat{v}_{1},\hat{v}_{2},...,\hat{v}_{N}\right)$,
where $\hat{v}_{n}\sim P\left(v_{n}\mid\hat{\bh}\right)$ and then
estimate the average error of $\bv$ and $\hat{\bv}$. When it comes
to the latter one, let $\bv_{\neg S}$ denote the set of observed
variables, and $\bv_{S}$ depict the set of unseen variables to be
fulfilled. It is necessary to compute the inference of $P\left(\bv_{S}\mid\bv_{\neg S}\right)$.
Since this inference demands an exponential computation if simultaneously
calculated, $P\left(v_{i}\mid\bv_{\neg S}\right)$, $i\in S$, is
estimated instead.

\subsubsection{Type-specific data of RBM\label{sub:RBM_Types}}

Recall that RBM is capable of modelling numerous types of data (e.g.,
binary, Gaussian, categorical, Poisson or count and softmax data).
For individual type, there are changes in the model and parameters
which causes to difference of energy, probabilities function. We are
assumed that hidden units are binary. Nonetheless, applying to variant
types of hidden units is absolutely similar to visible ones. 

Let $\tau\left(x\right)$ denote the logistic sigmoid function$\tau\left(x\right)=\frac{1}{1+e^{-x}}$.
The activation probability of hidden units are specified as $P\left(h_{j}=1\mid\bv\right)=\tau\left(b_{j}+\sum_{i}v_{i}W_{ij}\right)$.
Here we specify formula modifications for almost types of visible
units that RBM is able to handle.
\begin{itemize}
\item Binary data: $P\left(v_{i}=1\mid\bh\right)=\tau\left(a_{i}+\sum_{j}h_{j}W_{ij}\right)$.
\item Gaussian data: The logistic units have a poor representation for real-valued
data such as natural images, Mel-Cepstrum coefficients used to represent
speech \cite{mohamed_etal_nips09_dbn_pr}. These binary visible units
can be replaced by linear units with independent Gaussian noise. Let
$\sigma_{i}$ denote the standard deviation of the Gaussian noise
for visible unit $v_{i}$. The energy function then becomes

\begin{equation}
E\left(\bv,\bh\right)=\sum_{i=1}^{N}\frac{\left(v_{i}-a_{i}\right)^{2}}{2\sigma_{i}^{2}}-\sum_{j=1}^{K}b_{j}h_{j}-\sum_{i=1}^{N}\sum_{j=1}^{K}W_{ij}\frac{v_{i}}{\sigma_{i}}h_{j}\label{eq:RBM_Gaussian_Energy}
\end{equation}

where $\sigma_{i}$ is the standard deviation of the Gaussian noise
for visible unit $i$. The visible and hidden probabilities are computed
as follows

\begin{eqnarray*}
P\left(v_{i}=n\mid\bh\right) & = & \mathcal{N}\left(a_{i}+\sigma_{i}\sum_{j}h_{j}W_{ij};\sigma_{i}^{2}\right)\\
 & = & \frac{1}{\sqrt{2\pi}\sigma_{i}}\exp\left\{ -\frac{\left(n-a_{i}-\sigma_{i}\sum_{j}h_{j}W_{ij}\right)^{2}}{2\sigma_{i}^{2}}\right\} \\
P\left(h_{j}=1\mid\bv\right) & = & \tau\left(b_{j}+\sum_{i}W_{ij}\frac{v_{i}}{\sigma_{i}}\right)
\end{eqnarray*}

It is feasible to learn the variance of the noise for each visible
unit. However, it is complicated for using CD. To make learning easier,
we first normalise each feature of data to have zero mean and unit
variance. The standard variance $\sigma$ in Eq.~(\ref{eq:RBM_Gaussian_Energy})
is then set to 1. Then we can use noise-free top-down reconstructions
from hidden units to input data. The learning rate $\lambda$ in Eq.~(\ref{eq:RBM_paramUpdate})
must be smaller because there is no upper bound to the value.

\item Poisson data: To represent \emph{count} (e.g., document words, diagnosis
codes, bag-of-visual words), we adopt the constrained Poisson model
by \cite{salakhutdinov_hinton_SIGIR07_semantic}. Denote the Poisson
probability mass function by $Ps\left(n,\mu\right)=e^{-\mu}\frac{\mu^{n}}{n!}$,
in which $\mu$ is the mean or bias of the conditional Poisson model.
Visible probabilities are given by

\begin{eqnarray*}
P\left(v_{i}=n\mid\bh\right) & = & Ps\left(n,\frac{\exp\left(a_{i}+\sum_{j}h_{j}W_{ij}\right)}{\sum_{k}\exp\left(a_{k}+\sum_{j}h_{j}W_{ij}\right)}N\right)
\end{eqnarray*}

where $N=\sum_{i}v_{i}$ is the length of count data (e.g., document,
total number of diagnosis codes).

\item Softmax data: Referring to the binary data, the probability of turning
on a unit is given by the logistic sigmoid function. Each probability
of two possible states, i.e., 1 is on, 0 is off, is proportional to
the negative exponential of its energy. We can generalise to $M$
alternative states

\[
P\left(v_{i}^{m}=1\mid\bh\right)=\frac{\exp\left\{ a_{i}^{k}+\sum_{j}h_{j}W_{ij}^{m}\right\} }{\sum_{t}\exp\left\{ a_{i}^{t}+\sum_{j}h_{j}W_{ij}^{t}\right\} }
\]

These types of units can be called as ``softmax'' units. It is suitable
for modelling the data that has $M$ unordered options. As can be
seen, a softmax can be considered as a set of binary units. Thus each
binary unit in a softmax should be learned in the same way as standard
binary units. However, the hidden and visible probabilities are computed
differently. The hidden probability is now computed as

\[
P\left(h_{j}=1\mid\bv\right)=\tau\left(b_{j}+\sum_{i}\sum_{m}v_{i}^{m}W_{ij}^{m}\right)
\]

\textbf{\emph{Replicated Softmax}}:

When softmax units appear repeatedly in visible vectors, e.g. words
in documents, diagnosis codes in history of patients, Salakhutdinov~\emph{et~al}.
\cite{salakhutdinov_hinton_nips09_repsoftmax} introduced a separate
RBM with the number of softmax units be exactly the same as number
of units of a visible vector. These visible units can be called as
Replicated Softmax units. Assume that the order of units can be ignored,
entire softmax units can share the same set of weights as well as
parameters which connects them to hidden units. Let $\bv\in\left\{ q_{1},q_{2},..,q_{N}\right\} ^{D}$,
in which $N$ is total number of distinguished values that can be
assigned to one unit, $D$ is the actual number of units in a data
vector. Denote $K$ be the number of binary hidden units. Replicated
Softmax RBM models the appearance of visible softmax unit with $v_{i}^{n}=1$
if visible unit takes on $q_{n}$. Consider $\hat{v}_{i}=\sum_{d=1}^{D}v_{d}^{i}$
as the count for value $q_{i}$. The energy function is given as below

\begin{equation}
E\left(\bv,\bh\right)=-\sum_{i=1}^{N}a_{i}\hat{v}_{i}-D\sum_{j=1}^{K}b_{j}h_{j}-\sum_{i=1}^{N}\sum_{j=1}^{K}W_{ij}\hat{v}_{i}h_{j}\label{eq:RBM_RepSoftmax_Energy}
\end{equation}

As the same as softmax units, we have the visible and hidden probabilities

\begin{eqnarray}
P\left(v_{i}^{n}=1\mid\bh\right) & = & \frac{\exp\left\{ a_{n}+\sum_{j=1}^{K}h_{j}W_{nj}\right\} }{\sum_{t=1}^{N}\exp\left\{ a_{t}+\sum_{j=1}^{K}h_{j}W_{tj}\right\} }\label{eq:RBM_RepSoftmax_VisibleProbs}\\
P\left(h_{j}=1\mid\hat{\bv}\right) & = & \tau\left(Db_{j}+\sum_{i=1}^{N}W_{ij}\hat{v}_{i}\right)\label{eq:RBM_RepSoftmax_HiddenProbs}
\end{eqnarray}

\item Categorical data: Let $M$ describe the number of categories. Denote
by $S=\left\{ c_{1},c_{2},...,c_{M}\right\} $ the set of categories.
Let $\pi\left[v_{i}\right]$ be a $M$-dimension vector with $\pi_{m}\left[v_{i}\right]=1$
if $v_{i}=c_{m}$ and 0 otherwise. Here are the following energy,
visible and hidden probabilities functions

\begin{eqnarray*}
E\left(\bv,\bh\right) & = & -\sum_{i=1}^{N}\sum_{m=1}^{M}a_{i}^{m}\pi_{m}\left[v_{i}\right]-\sum_{j=1}^{K}b_{j}h_{j}-\sum_{i=1}^{N}\sum_{m=1}^{M}\sum_{j=1}^{K}W_{ij}^{m}\pi_{m}\left[v_{i}\right]h_{j}\\
P\left(v_{i}^{m}=1\mid\bh\right) & = & \frac{\exp\left\{ \sum_{m=1}^{M}a_{i}^{m}\pi_{m}\left[v_{i}\right]+\sum_{m=1}^{M}\sum_{j=1}^{K}W_{ij}^{m}\pi_{m}\left[v_{i}\right]h_{j}\right\} }{\sum_{l=1}^{M}\exp\left\{ a_{i}^{l}+\sum_{j=1}^{K}W_{ij}^{l}h_{j}\right\} }\\
P\left(h_{j}=1\mid\bv\right) & = & \tau\left(b_{j}+\sum_{i=1}^{N}\sum_{m=1}^{M}W_{ij}^{m}\pi_{m}\left[v_{i}\right]\right)
\end{eqnarray*}

\end{itemize}

\subsection{Mixed-Variate Restricted Boltzmann Machine\label{sub:MV.RBM}}

\subsubsection{Model representation}

Mixed-Variate Restricted Boltzmann Machine (MV.RBM) is introduced
by Truyen~\emph{et~al.} \cite{truyen_phung_venkatesh_jmlr11_mvrbm}.
A Mixed-Variate RBM is a RBM with heterogeneous input units, each
of which has the own type. See, for example, Figure~\ref{fig:MVRBMs_patient_data}
in Section~\ref{sub:MDA_Framework} for an illustration. More formally,
let $\bv$ denote the joint set of visible variables: $\bv=\left(v_{1},v_{2},...,v_{N}\right)$,
$\bh$ the joint set of binary hidden units: $\bh=\left(h_{1},h_{2},...,h_{K}\right)$,
where $h_{k}\in\left\{ 0,1\right\} $ for all $k$. Each visible unit
encodes type-specific information, and the hidden units capture the
\emph{latent factors} not presented in the observations. Thus the
MV.RBM can be seen as a way to transform inhomogeneous observational
record into a \emph{homogeneous representation} of the patient profile.
Another way to view this as a mixture model where there are $2^{K}$
mixture components. This capacity is arguably important to capture
all factors of variation in the patient cohort.

The MV.RBM defines a Boltzmann distribution over all variables: $P\left(\bv,\bh;\psi\right)=e^{-E\left(\bv,\bh\right)}/Z\left(\psi\right)$,
where $E\left(\bv,\bh\right)$ is model energy, $Z\left(\psi\right)$
is the normalising constant and $\psi$ is model parameter. In particular,
the energy is defined as{\small{}
\begin{equation}
E\left(\bv,\bh;\psi\right)=-\left(\sum_{i}F(v_{i})+\boldsymbol{a}^{\top}\bv+\boldsymbol{b}^{\top}\bh+\bh^{\top}\boldsymbol{W}\bv\right)\label{eq:MVRBM_Energy}
\end{equation}
}{\small \par}

\noindent where $F(v_{i})$ is the type-specific function, $\boldsymbol{a}=\left(a_{1},a_{2},...,a_{N}\right)$
and $\boldsymbol{b}=\left(b_{1},b_{2},...,b_{K}\right)$ are biases
of visible and hidden units respectively, and $\boldsymbol{W}=\left[W_{ij}\right]$
are the weights connecting hidden and visible units. Because the visible
variables of MV.RBM is mixed-variate, $\bv$ consists of $\left(\bv_{disc},\bv_{cont}\right)$,
where $\bv_{disc}$ defines the joint set of discrete variables and
$\bv_{cont}$ is the joint set of continuous ones. The integration
must be involved into the partition function in Eq.~(\ref{eq:RBM_Part}),
which is given by

\[
\mathcal{Z}\left(\psi\right)=\int_{\bv_{cont}}\left(\sum_{\bv_{dist},\bh}\exp\left\{ -E\left(\bv_{dist},\bv_{cont},\bh;\psi\right)\right\} \right)d\left(\bv_{cont}\right)
\]

\noindent Similarity to plain RBM, the MV.RBM comprises conditional
independence among intra-layer variables which lead to hidden and
visible factorisations in Eqs.~(\ref{eq:RBM_factorVis},\ref{eq:RBM_factorHid}).
The type-specific function $F(v_{i})$ can be pointed out for individual
type basing on its probabilistic function in Section~\ref{sub:RBM_Types}.
For example, For example, let $f_{i}(\bh)=a_{i}+\sum_{k}W_{ik}h_{k}$,
the \emph{binary} units would be specified as: $P\left(v_{i}\mid\bh\right)=1/\left(1+e^{-f_{i}(\bh)}\right)$
(i.e., $F_{i}(v_{i})=0$); the \emph{Gaussian} units: $P\left(v_{i}\mid\bh\right)=\mathcal{N}\left(\sigma_{i}^{2}f_{i}(\bh);\sigma_{i}\right)$
(i.e., $F_{i}(v_{i})=-v_{i}^{2}/2\sigma_{i}^{2}$), and the \emph{categorical}
units: $P\left(v_{i}\mid\bh\right)=e^{f_{i}(\bh)}/\sum_{j}e^{f_{j}(\bh)}$.

\subsubsection{Extensions\label{sub:MVRBM_Ext}}

The Poisson and ``replicated softmax'' are not readily available
in the current MV.RBM of Truyen~\emph{et~al.} \cite{truyen_phung_venkatesh_jmlr11_mvrbm}.
In this work, we attempt to introduce these two types to the machine.
To be easier, we consider visible units represent words in a document
for an instance. The other count data (e.g., bag-of-visual words,
diagnosis codes of a patient) are equivalent. To represent \emph{counts},
we adopt the constrained Poisson model by \cite{salakhutdinov_hinton_SIGIR07_semantic}
in that $P\left(v_{i}=n\mid\bh\right)=\mbox{Poisson}\left(n,\frac{\exp\left\{ f_{i}(\bh)\right\} }{\sum_{k}\exp\left\{ f_{i}(\bh)\right\} }L\right)$,
where $L$ is the ``document'' length. With ``replicated softmax'',
we adopt the idea from \cite{salakhutdinov_hinton_nips09_repsoftmax}
where repeated words share the same parameters. In the end, we build
one MV.RBM per document due to the difference in the word sets. Further,
to balance the contribution of the hidden units against the variation
in input length, it is important to make a change to the energy model
in Eq.~(\ref{eq:MVRBM_Energy}) as follows: $D\bb\leftarrow\bb$
where $D$ is the total number of input variables for each document.
We show the efficiency of these parameter sharing and balancing in
Section~\ref{sec:apps}.

\subsubsection{Parameters learning and Inference}

Because the gradient of log-likelihood still forms model and data
expectations as in Eq.~(\ref{eq:RBM_logLLGrad}), Contrastive Divergence
(CD) obviously can be used to approximate model expectation. Then
we apply gradient ascent manner as in Eq.~(\ref{eq:RBM_paramUpdate})
to learn parameters of MV.RBM.

In \cite{truyen_phung_venkatesh_jmlr11_mvrbm}, Truyen~\emph{et~al}.
also show that the MV.RBM is capable to construct a predictive model
and then do prediction on unseen variables given observed ones. Let
$\bv_{S}$ depict the set of unseen variables to be fulfilled,$\bv_{\neg S}$
denote the set of observed variables. The problem is of estimating
the conditional distribution $P\left(\bv_{S}\mid\bv_{\neg S}\right)$.
Assuming that unseen variables are pair-wise independent, $P\left(v_{i}\mid\bv_{\neg S}\right)$,
$i\in S$, is estimated instead. There are two common ways: \emph{generative}
and \emph{discriminative} methods to learn a predictive model. The
former attempts to model the joint distribution $P\left(v_{i},\bv_{\neg S}\right)$.
The learning is now of maximising the following likelihood

\begin{equation}
\likelihood_{1}=\sum_{\bv_{\neg S}}\tilde{P}\left(\bv_{\neg S}\right)\log P\left(\bv_{\neg S}\right)\label{eq:MVRBM_PredL1}
\end{equation}

\noindent The latter one is to directly model the conditional distribution
$P\left(v_{i}\mid\bv_{\neg S}\right)$ which is equivalent to maximise
the conditional likelihood given as below

\begin{equation}
\likelihood_{2}=\sum_{v_{i}}\sum_{\bv_{\neg S}}\tilde{P}\left(v_{i},\bv_{\neg S}\right)\log P\left(v_{i}\mid\bv_{\neg S}\right)\label{eq:MVRBM_Pred2}
\end{equation}

\noindent It can be seen that the discriminative method requires less
computational complexity since we do not have to estimate $P\left(\bv_{\neg S}\right)$.
But the capability of MV.RBM is to unsupervised learn joint distribution
and then do projections $P\left(\bh\mid\bv_{\neg S}\right)$, in the
work of Truyen~\emph{et~al}., they follow the \emph{third} approach.
It is a \emph{hybrid} method that linear combine generative and discriminative
methods as follows

\begin{eqnarray*}
\likelihood_{3} & = & \lambda\likelihood_{1}+\left(1-\lambda\right)\likelihood_{2}\\
 & = & \lambda\sum_{\bv_{\neg S}}\tilde{P}\left(\bv_{\neg S}\right)\log P\left(\bv_{\neg S}\right)+\left(1-\lambda\right)\sum_{v_{i}}\sum_{\bv_{\neg S}}\tilde{P}\left(v_{i},\bv_{\neg S}\right)\log P\left(v_{i}\mid\bv_{\neg S}\right)
\end{eqnarray*}

\noindent where $\lambda\in\left(0,1\right)$ is the coefficient to
adjust the effect of two methods on the final likelihood. There are
two ways to maximise the likelihood $\likelihood_{3}$. One optimises
both likelihood $\likelihood_{1}$ and $\likelihood_{2}$ simultaneously.
Another is to use 2-stage procedure: first pretrain $P\left(\bv_{\neg S}\right)$
using unsupervised fashion and then fine-tune the predictive model
$P\left(v_{i}\mid\bv_{\neg S}\right)$ by discriminative training.

\subsection{Integrating Structured Sparsity\label{sub:Sparsity}}

Recall that hidden posteriors which are projected using RBM or MV.RBM
form latent representations. These latent representations capture
the regularities of the data. Nonetheless, they are not readily unstructured
and disentangle all the factors of variation. Since, for example,
the image frequently describes a few types of objects, the probabilistic
activations of hidden units should be controlled for better representation.
One way to do improve representation is to incorporate structured
sparsity which supports better separation of object groups and easier
interpretation. Following the previous work \cite{luo_etal_aaai11_sgrbm},
we apply mixed-norm $\nicefrac{l_{1}}{l_{2}}$ regulariser on the
latent representation. Hidden units in latent representation are also
equally arranged into non-overlapped groups. The mixed-norm regulariser
can control both inter-group and intra-group hidden activations.

Denote $\mathcal{H}$ by the set of all hidden units indices: $\mathcal{H}=\left\{ 1,2,...,K\right\} $.
We divide hidden units into $M$ non-overlapped groups with equal
sizes. Let $\mathcal{G}_{m}$ describe the indices of hidden units
in group $m^{th}$, where $G_{m}\subset\mathcal{H}$. $l_{2}\textnormal{-norm}$
of the $m^{th}$ group is given by

{\small{}
\begin{equation}
\mathcal{R}{}_{2}^{\left(m\right)}\left(\bv\right)=\sqrt{\sum_{j\in G_{m}}P\left(h_{j}=1\mid\bv\right)^{2}}\label{eq:SGRBM_L2_norm}
\end{equation}
}{\small \par}

From that, the mixed-norm $\nicefrac{l_{1}}{l_{2}}$ is defined as
$l_{1}\textnormal{-norm}$ regularising over all $M$ groups as follows

{\small{}
\begin{equation}
\mathcal{R}\left(\bv\right)=\sum_{m=1}^{M}\left|\mathcal{R}_{2}^{\left(m\right)}\left(\bv\right)\right|=\sum_{m=1}^{M}\sqrt{\sum_{j\in G_{m}}P\left(h_{j}=1\mid\bv\right)^{2}}\label{eq:SGRBM_L1_norm}
\end{equation}
}{\small \par}

The way to integrate structured sparsity is to involve the regulariser
into the log-likelihood of RBM. The learning of RBM then operates
both updating parameters and enhancing the structured sparsity of
latent representations simultaneously. The integration defines a new
likelihood function of RBM as below

\begin{equation}
\mathcal{L}_{spr}\left(\bv;\psi\right)=\log\mathcal{L}\left(\bv;\psi\right)-\alpha\mathcal{R}\left(\bv\right)\label{eq:SGRBM_Likelihood}
\end{equation}

\noindent where $\alpha\geq0$ is the regularising constant for sparsity.
Basing on the gradient of log-likelihood function in Eq.~(\ref{eq:RBM_logLLGrad}),
we obtain the following gradient of the likelihood 

\begin{equation}
\frac{\partial}{\partial\psi}\mathcal{L}_{spr}\left(\bv;\psi\right)=\E_{\bv,\bh}\left[\frac{\partial E\left(\bv,\bh\right)}{\partial\psi}\right]-\E_{\bh\mid\bv}\left[\frac{\partial E\left(\bv,\bh\right)}{\partial\psi}\right]-\alpha\frac{\partial}{\partial\psi}\mathcal{R}\left(\bv\right)\label{eq:SGRBM_logLLGrad}
\end{equation}

The derivatives of first two expectations in Eq.~(\ref{eq:SGRBM_logLLGrad})
with respect to model parameters can be estimated using Eqs.~(\ref{eq:paramsWGrad},\ref{eq:paramsAGrad},\ref{eq:paramsBGrad}).
Here we specify the added regularisation $\mathcal{R}\left(\bv\right)$.
Note that $\left|\mathcal{R}_{m}\left(\bv\right)\right|=\mathcal{R}_{m}\left(\bv\right)$
since $\mathcal{R}_{m}\left(\bv\right)\geq0$. Thus, we have these
following deductions

\begin{eqnarray*}
\mathcal{R}\left(\bv\right) & = & \sum_{m=1}^{M}\mathcal{R}_{m}\left(\bv\right)\\
 & = & \sum_{m=1}^{M}\sqrt{\sum_{t\in\mathcal{G}_{m}}P\left(h_{t}=1\mid\bv\right)^{2}}\\
 & = & \sqrt{\sum_{t\in\mathcal{G}_{\hat{m}}}P\left(h_{t}=1\mid\bv\right)^{2}}\\
 & = & \sqrt{\sum_{t\in\mathcal{G}_{\hat{m}},t\neq j}P\left(h_{t}=1\mid\bv\right)^{2}+P\left(h_{j}=1\mid\bv\right)^{2}}
\end{eqnarray*}

Taking the derivative of the regularisation with respect to inter-layer
parameter $W_{ij}$ and the hidden bias $b_{j}$, it reads

\begin{eqnarray*}
\frac{\partial}{\partial b_{j}}\mathcal{R}\left(\bv\right) & = & \frac{P\left(h_{j}=1\mid\bv\right)}{\sqrt{\sum_{t\in\mathcal{G}_{\hat{m}}}P\left(h_{t}=1\mid\bv\right)^{2}}}\frac{\partial}{\partial b_{j}}P\left(h_{j}=1\mid\bv\right)\\
 & = & \frac{P\left(h_{j}=1\mid\bv\right)^{2}P\left(h_{j}=0\mid\bv\right)}{\sqrt{\sum_{t\in\mathcal{G}_{\hat{m}}}P\left(h_{t}=1\mid\bv\right)^{2}}}\\
\frac{\partial}{\partial W_{ij}}\mathcal{R}\left(\bv\right) & = & \frac{\partial}{\partial b_{j}}\mathcal{R}\left(\bv\right)v_{i}
\end{eqnarray*}

During maximising the likelihood, this regulariser is minimised and
this leads to group-wise sparsity, i.e., only few groups of hidden
units will be activated (e.g., refer to the last column of Figure~\ref{fig:IR_EgImg_NUS-WIDE-animal.}
in Section~\ref{sub:IR}). The parameters are updated using stochastic
gradient ascent fashion as follows

\[
\psi\leftarrow\psi+\lambda\left(\frac{\partial}{\partial\psi}\mathcal{L}_{spr}\right)
\]

\noindent for some learning rate $\alpha>0$.

\subsection{Learning Distance Metric\label{sub:metric_learning}}

Latent representation may not fully capture \emph{intra/inter-concept
variation}s, and thus it may not result in a good distance metric
for retrieval tasks, e.g., image retrieval. It is better to directly
learn a distance metric that suppresses intra-concept variation and
enlarges inter-concept variation. Given the probabilistic nature of
our representation, a suitable distance is the symmetric Kullback-Leibler
divergence, also known as Jensen-Shannon divergence:

\begin{eqnarray}
\dataset\left(g,f\right) & = & \frac{1}{2}\left(\textnormal{KL}\left(g\Vert f\right)+\textnormal{KL}\left(f\Vert g\right)\right)\label{eq:MLRBM_JenShanDiv}
\end{eqnarray}

\noindent where $\textnormal{KL}\left(g\Vert f\right)=\sum_{\bh}P\left(\bh\mid g\right)\log\frac{P\left(\bh\mid g\right)}{P\left(\bh\mid f\right)}$.
Let $\mathsf{N}\left(f\right)$ denote the set of other objects that
share the same concept with the object $f$, and $\bar{\mathsf{N}}\left(f\right)$
denotes those do not. The mean distance to all other objects in $\mathsf{N}\left(f\right)$
should be minimised

\begin{equation}
\dataset_{\mathsf{N}\left(f\right)}=\frac{1}{\left|\mathsf{N}\left(f\right)\right|}\sum_{g\in\mathsf{N}\left(f\right)}\dataset\left(P\left(\bh\mid\bv^{\left(g\right)}\right),P\left(\bh\mid\bv^{\left(f\right)}\right)\right)\label{eq:MLRBM_NBdist}
\end{equation}
On the other hand, the mean distance to all images in $\bar{\mathsf{N}}\left(f\right)$
should be enlarged

\begin{equation}
\dataset_{\bar{\mathsf{N}}\left(f\right)}=\frac{1}{\left|\bar{\mathsf{N}}\left(f\right)\right|}\sum_{g\in\bar{\mathsf{N}}\left(f\right)}\dataset\left(P\left(\bh\mid\bv^{\left(g\right)}\right),P\left(\bh\mid\bv^{\left(f\right)}\right)\right)\label{eq:MLRBM_NNBdist}
\end{equation}

\noindent The idea of intra-concept distance has been studied in \cite{truyen_phung_venkatesh_icme12_mlrbm}.
We here enhance the effect of distance metric on learning latent representation
with the inter-concept distance. The log-likehood transforms into
the new likelihood as follows

\[
\likelihood_{ml}=\sum_{f}\log\mathcal{L}\left(\bv^{\left(f\right)};\psi\right)-\beta\left(\sum_{f}\dataset_{\mathsf{N}\left(f\right)}-\sum_{f}\dataset_{\bar{\mathsf{N}}\left(f\right)}\right)
\]

\noindent where $\beta\geq0$ is the coefficient to control the effect
of distance metrics. Maximising this likelihood is now equivalent
to simultaneously maximising the data likelihood $\mathcal{L}\left(\bv;\psi\right)$,
minimising the neighbourhood distance $\dataset_{\mathsf{N}\left(f\right)}$
and maximising the non-neighbourhood distance $\dataset_{\bar{\mathsf{N}}\left(f\right)}$.
From Eq.~(\ref{eq:RBM_logLLGrad}), the gradient of the new likelihood
can be given by

\begin{equation}
\frac{\partial}{\partial\psi}\mathcal{L}_{ml}\left(\bv;\psi\right)=\E_{\bv,\bh}\left[\frac{\partial E\left(\bv,\bh\right)}{\partial\psi}\right]-\E_{\bh\mid\bv}\left[\frac{\partial E\left(\bv,\bh\right)}{\partial\psi}\right]-\beta\frac{\partial}{\partial\psi}\left(\sum_{f}\dataset_{\mathsf{N}\left(f\right)}-\sum_{f}\dataset_{\bar{\mathsf{N}}\left(f\right)}\right)\label{eq:MLRBM_logLLGrad}
\end{equation}

\noindent In Eqs.~(\ref{eq:paramsWGrad},\ref{eq:paramsAGrad},\ref{eq:paramsBGrad})
above, we already can estimate the derivatives of two expectations.
To compute the gradient of the mean distances $\dataset_{\mathsf{N}\left(f\right)}$
and $\dataset_{\bar{\mathsf{N}}\left(f\right)}$ defined in Eqs.(\ref{eq:MLRBM_NBdist},\ref{eq:MLRBM_NNBdist}),
we need the gradient for each pairwise distance $\dataset\left(g,f\right)=\dataset\left(P\left(\bh\mid\bv^{\left(g\right)}\right),P\left(\bh\mid\bv^{\left(f\right)}\right)\right)$.
Taking derivative of the metric distance function with respect to
parameter $\psi_{\bullet j}$, we have

\begin{equation}
\frac{\partial\dataset\left(g,f\right)}{\partial\psi_{\bullet j}}=\frac{\partial\dataset\left(g,f\right)}{\partial P\left(h_{j}^{1}\mid f\right)}\frac{\partial P\left(h_{j}^{1}\mid f\right)}{\partial\psi_{\bullet j}}+\frac{\partial\dataset\left(g,f\right)}{\partial P\left(h_{j}^{1}\mid g\right)}\frac{\partial P\left(h_{j}^{1}\mid g\right)}{\partial\psi_{\bullet j}}\label{eq:MLRBM_GradML}
\end{equation}

\noindent in which $P\left(h_{j}^{1}\mid f\right)$ is the shorthand
for $P\left(h_{j}=1\mid\bv^{\left(f\right)}\right)$, $\psi_{\bullet j}$
is the parameter associated with hidden units $h_{j}$. Hidden units
are assumed to be binary units. Recall from Section~\ref{sub:RBM_Types}
that the probabilistic activations of hidden units are sigmoid functions.
Thus the partial derivatives with respect to the mapping column $W_{\bullet j}$
and bias $b_{j}$ are then

\noindent 
\begin{eqnarray}
\frac{\partial P\left(h_{j}^{1}\mid f\right)}{\partial W_{\bullet j}} & = & P\left(h_{j}^{1}\mid f\right)\left(1-P\left(h_{j}^{1}\mid f\right)\right)\bv^{\left(f\right)}\label{eq:MLRBM_GradPw}\\
\frac{\partial P\left(h_{j}^{1}\mid f\right)}{\partial b_{j}} & = & P\left(h_{j}^{1}\mid f\right)\left(1-P\left(h_{j}^{1}\mid f\right)\right)\label{eq:MLRBM_GradPb}
\end{eqnarray}

\noindent As defined in Eq.~(\ref{eq:MLRBM_JenShanDiv}), the derivatives
$\frac{\partial\dataset\left(g,f\right)}{\partial P\left(h_{j}^{1}\mid f\right)}$
and $\frac{\partial\dataset\left(g,f\right)}{\partial P\left(h_{j}^{1}\mid g\right)}$
depend on the derivative of the KL-divergence, which reads

\begin{eqnarray}
\frac{\partial\textnormal{KL}\left(g\Vert f\right)}{\partial P\left(h_{j}^{1}\mid g\right)} & = & \sum_{j}\left(\log\frac{P\left(h_{j}^{1}\mid g\right)}{P\left(h_{j}^{1}\mid f\right)}-\log\frac{1-P\left(h_{j}^{1}\mid g\right)}{1-P\left(h_{j}^{1}\mid f\right)}\right)\label{eq:MLRBM_KLDivGFGradG}\\
\frac{\partial\textnormal{KL}\left(f\Vert g\right)}{\partial P\left(h_{j}^{1}\mid g\right)} & = & \sum_{j}\left(-\frac{P\left(h_{j}^{1}\mid f\right)}{P\left(h_{j}^{1}\mid g\right)}+\frac{1-P\left(h_{j}^{1}\mid f\right)}{1-P\left(h_{j}^{1}\mid g\right)}\right)\label{eq:MLRBM_KLDivFGGradG}
\end{eqnarray}

\noindent Equivalently to Eqs.~(\ref{eq:MLRBM_KLDivGFGradG},\ref{eq:MLRBM_KLDivFGGradG}),
$\frac{\partial\textnormal{KL}\left(g\Vert f\right)}{\partial P\left(h_{j}^{1}\mid f\right)}$
and $\frac{\partial\textnormal{KL}\left(f\Vert g\right)}{\partial P\left(h_{j}^{1}\mid f\right)}$
can be easily computed. Combining all things in Eqs.~(\ref{eq:MLRBM_GradPw},\ref{eq:MLRBM_GradPb},\ref{eq:MLRBM_KLDivGFGradG},\ref{eq:MLRBM_KLDivFGGradG}),
we can calculate the derivative of the metric distance with respect
to parameter $\psi_{\bullet j}$ in Eq.~(\ref{eq:MLRBM_GradML}).

Finally, the parameters are updated using stochastic gradient ascent
fashion as follows

\[
\psi\leftarrow\psi+\lambda\left(\frac{\partial}{\partial\psi}\mathcal{L}_{ml}\right)
\]

\noindent for some learning rate $\alpha>0$.

\section{Applications\label{sec:apps}}

\noindent In this section, we demonstrate the performances of our
proposed models in various applications including \emph{latent patient
profile} modelling in medical data analysis (published in \cite{tu_etal_pakdd13_latent})
and representation learning for image retrieval (published in \cite{tu_etal_icme13_learning}).
The former application is to implement an extension of Mixed-Variate
Restricted Boltzmann Machine (MV.RBM) for clustering patients and
predicting diagnosis codes $1$-year in advance. The latter one learns
sparse latent representation and distance metric for image retrieval.
The experimental results demonstrate the former model gains better
results than baseline methods and the latter outperforms state-of-the-art
rivals on NUS-WIDE datasets \cite{chua_etal_civr09_nuswide}.

\subsection{Medical data analysis}

\noindent \textcolor{red}{}It is undeniable that health is the most
significant factor of human being. Unfortunately, the modern world
has been seeing more and more emerging incurable diseases. They adversely
impact human life quality as well as consume a huge amount of healthcare
expenditures \cite{bodenheimer_etal_ha09_confronting}. Besides, chronic
illness often extends over many years along with acute exacerbations,
progressive deterioration and complications. For example, \emph{Diabetes
Mellitus} is one such chronic disease from which $346$ million people
worldwide suffer, as estimated by The World Health Organization (WHO)
\cite{WHO_Diabetes}. Only about $5-10$\% of them have Type I diabetes
mellitus, whilst Type II comprises the rest. The people who suffer
Type I are not able to produce insulin. In contrast, Type II diabetes
means that there is an inability to absorb insulin. In 2004, $3.4$
million people died from complications of high blood sugar. The incidence
of diabetes mellitus is increasing, and being diagnosed in younger
people. This leads to serious complications associated with long disease
duration - deterioration in blood vessels, eyes, kidneys and nerves.
It is a chronic, lifelong disease.

Consequently, healthcare management requires a scaled-up investigation
to improve chronic illness treatment and minimise the exacerbations,
slow or halt deterioration, and decrease the cost of care. To provide
high quality healthcare, care plans are issued to patients to manage
them within the community, taking steps in advance so that these people
are not hospitalised. For instance, identifying groups of patients
with similar characteristics help them be covered by a coherent care
plan. Additionally, if the hospital can predict the disease codes
arising from escalating complication of chronic disease, it can adjust
financial and manpower resources. Thus useful prediction of codes
for chronic disease can lead to service efficiency. Thanks to recent
high-tech equipment and information technology applied in healthcare
organisations, the quantity of electronic hospital records are substantially
increasing. Effective patterns and latent structures discovery from
this data can improve healthcare system and optimise the pathology
treatment for patients. It does require a rigorous analysis on the
data.

Clustering is a natural selection for this task. However, medical
data is complex -- it is mixed-type containing Boolean data (e.g.,
male/female), continuous quantities (e.g., age), single categories
(e.g., regions), and repeated categories (e.g., disease codes). Traditional
clustering methods cannot naturally integrate such data and we choose
the extension (see Section~\ref{sub:MVRBM_Ext}) of the MV.RBM \cite{truyen_phung_venkatesh_jmlr11_mvrbm}.
The Mixed-Variate RBM uncovers \emph{latent profile} factors, enabling
subsequent clustering. Using a cohort of $6,931$ chronic diabetes
patients with data from 2007 to 2011, we collect $3,178$ diagnosis
codes (treated as repeated categories) and combine it with region-of-birth
(as categories) and age (as Gaussian variables) to form our dataset.
We show clustering results obtained from running affinity propagation
(AP) \cite{frey_dueck_sci07_ap}, containing 10 clusters and qualitatively
evaluate the disease codes of groups. We demonstrate that the Mixed-Variate
RBM followed by AP outperforms all baseline methods -- Bayesian mixture
model and affinity propagation on the original diagnosis codes, and
$k$-means and AP on latent profiles, discovered by just the plain
RBM \cite{salakhutdinov_hinton_nips09_repsoftmax}. Predicting disease
codes for future years enables hospitals to prepare finance, equipment
and logistics for individual requirements of patients. Thus prediction
of disease codes forms the next part of our study. Using the Mixed-Variate
RBM and the dataset described above, we demonstrate that our method
outperforms other methods, establishing the versatility of the latent
profile discovery with MV.RBM.

The next section presents our patient profile modelling framework.
Then, we describe our implementation on the diabetes cohort and demonstrate
efficiencies of our methods.

\subsubsection{Latent Patient Profiling\label{sub:MDA_Framework}}

A patient profile in modern hospitals typically consists of multiple
records including demographics, admissions, diagnoses, surgeries,
pathologies and medication. Each record contains several fields, each
of which is type-specific. For example, age can be considered as a
continuous quantity but a diagnosis code is a discrete element. At
the first approximation, each patient can be represented by using
a long vector of mixed types\footnote{Since each field may be repeated over time (e.g., diagnosis codes),
we need an aggregation scheme to summarize the field. Here we use
the simple counting for diagnosis codes.}. However, joint modelling of mixed types is known to be highly challenging
even for a small set of variables \cite{mcculloch_smmr08_joint,dunson_herring_biostat05_bayesian}.
Complete patient profiling, on the other hand, requires handling of
thousands of variables. Of equal importance is that the profiling
should readily support a variety of clinical analysis tasks such as
patient clustering, visualisation and disease prediction. In what
follows, we develop a representational and computational scheme to
capture such heterogeneity in an efficient way.

\begin{figure}[t]
\begin{centering}
\includegraphics[width=0.6\textwidth]{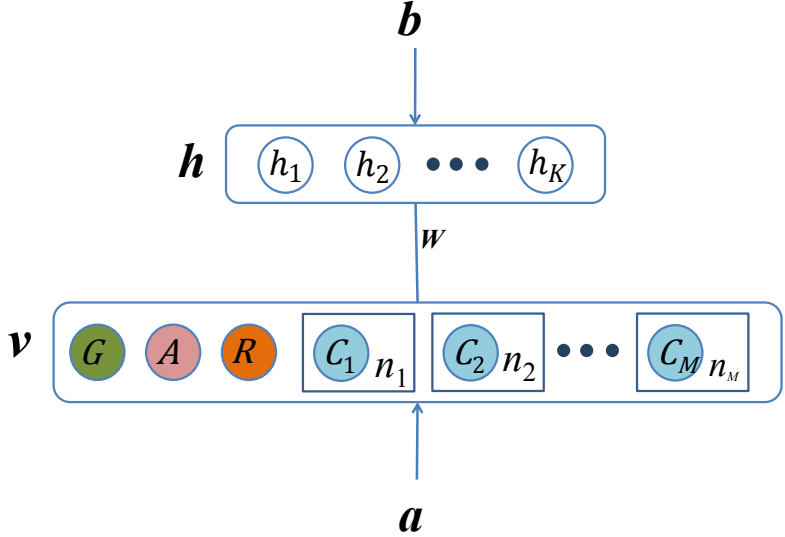}
\par\end{centering}

\caption{Patient profiling using Mixed-Variate RBMs. The top layer represents
stochastic binary units. The bottom layer encodes multiple type-specific
inputs: \emph{A} for age (continuous), \emph{G} for gender (binary),
\emph{R} for region-of-birth, $C_{k}$ for diagnosis codes. The circles
within squares denote the replicated diagnosis codes (categorical)
where the integers $\{n_{k}\}$ denotes the number of replications.\label{fig:MVRBMs_patient_data}}
\end{figure}

The goal of patient profiling is to construct an effective \emph{personal
representation} from multiple hospital records. Here we focus mainly
on patient demographics (e.g., \emph{age}, \emph{gender} and \emph{region-of-birth})
and their existing health conditions (e.g., existing \emph{diagnoses}).
For simplicity, we consider a binary gender (male/female). Further,
age can be considered as a continuous quantity and thus a Gaussian
unit can be used\footnote{Although the distribution of ages for a particular disease is generally
not Gaussian, our model is a mixture of exponentially many components
($2^{K}$, see Section~\ref{sub:RBM} for detail), and thus can capture
any distribution with high accuracy.}; and region-of-birth and diagnosis as categorical variables. However,
since the same diagnosis can be repeated during the course of readmissions,
it is better to include them all. In particular, we use the extension
of MV.RBM with repeated diagnoses considered as ``replicated softmax''
units (see Section~\ref{sub:MVRBM_Ext}). See Figure~\ref{fig:MVRBMs_patient_data}
for the illustration of MV.RBM for patient profiling.

Once the model has been estimated, the \emph{latent profiles} are
generated by computing the posterior vector $\hat{\bh}=\left(P\left(h_{1}^{1}\mid\bv\right),P\left(h_{2}^{1}\mid\bv\right),...,P\left(h_{K}^{1}\mid\bv\right)\right)$,
where $P\left(h_{k}^{1}\mid\bv\right)$is a shorthand for $P\left(h_{k}=1\mid\bv\right)$
-- the probability that the $k$-th latent factor is activated given
the demographic and clinical input $\bv$:
\[
P\left(h_{k}^{1}\mid\bv\right)=\frac{1}{1+\exp\left\{ -Db_{k}-\sum_{i}W_{ik}v_{i}\right\} }
\]

\noindent where $D$ is the total number of input variables for each
patient. As we will then demonstrate in Section~\ref{sub:MDA_Results},
the latent profile can be used as input for a variety of analysis
tasks such as patient clustering and visualisation.

Interestingly, the model also enables a certain degree of \emph{disease
prediction}, i.e., we want to guess which diagnoses will be positive
for the patient in the future\footnote{Although this appears to resemble the traditional collaborative filtering,
it is more complicated since diseases may be recurrent, and the strict
temporal orders must be observed to make the model clinically plausible.}. Although this may appear to be an impossible task, it is plausible
statistically because some diseases are highly correlated or even
causative, and there are certain pathways that a disease may progress.
More specifically, subset of diagnoses at time $t+1$ can be predicted
by searching for the mode of the following conditional distribution:
\[
P\left(\bv^{(t+1)}\mid\bv^{(1:t)}\right)=\sum_{\bh}P\left(\bv^{(t+1)},\bh\mid\bv^{(1:t)}\right)
\]

\noindent Unfortunately the search is intractable as we need to traverse
through the space of all possible disease combinations, which has
the size of $2^{M}$ where $M$ is the set of diagnosis codes. To
simplify the task and to reuse of the readily discovered latent profile
$\hat{\bh}^{(1:t)}$, we assume that (i) the model distribution is
not changed due to the ``unseen'' future, (ii) the latent profile
at this point captures everything we can say about the state of the
patient, and (iii) future diseases are conditionally independent given
the current latent profile. This leads to the following \emph{mean-field}
approximation\footnote{This result is obtained by first disconnecting the future diagnosis-codes
from the latent units and then find the \emph{suboptimal} factorised
distribution $Q\left(\bv^{(t+1)},\bh\mid\bv^{(1:t)}\right)=\prod_{j}Q_{j}\left(v_{j}^{(t+1)}\mid\bv^{(1:t)}\right)\prod_{k}P\left(h_{k}\mid\bv^{(1:t)}\right)$
that minimises the Kullback-Leibler divergence from the original distribution
$P\left(\bv^{(t+1)},\bh\mid\bv^{(1:t)}\right)$.}:
\begin{equation}
P\left(v_{j}^{(t+1)}\mid\bv^{(1:t)}\right)\approx\frac{\exp\left\{ a_{j}+\sum_{k}W_{jk}P\left(h_{k}^{1}\mid\bv^{(1:t)}\right)\right\} }{\sum_{i}\exp\left\{ a_{i}+\sum_{k}W_{ik}P\left(h_{k}^{1}\mid\bv^{(1:t)}\right)\right\} }\label{eq:MDA_diseasePrediction}
\end{equation}

\subsubsection{Implementation and Results\label{sub:MDA_Results}}

Here we present the analysis of patient profiles using the data obtained
from Barwon Health, Victoria, Australia\footnote{Ethics approval 12/83.},
during the period of $2007-2011$ using the extended MV.RBM described
in Section~\ref{sub:MDA_Framework}. In particular, we evaluate the
capacity of the MV.RBM for patient clustering and for predicting future
diseases. For the former task, the MV.RBM is can be seen as a way
to transform complex input data into a homogeneous vector from which
post-processing steps (e.g., clustering and visualisation) can take
place. For the prediction task, the MV.RBM acts as a classifier that
map inputs into outputs.

Our main interest is in the \emph{diabetes} cohort of $7,746$ patients.
There are two types of diabetes: Type I is typically present in younger
patients who are not able to produce insulin; and Type II is more
popular in the older group who, on the other hand, cannot adsorb insulin.
One of the most important indicators of diabetes is the high blood
sugar level compared to the general population. Diabetes are typically
associated with multiple diseases and complications: The cohort contains
$5,083$ distinct diagnosis codes, many of which are related to other
conditions and diseases such as obesity, tobacco use and heart problems.
For robustness, we remove those rare diagnosis codes with less than
$4$ occurrences in the data. This results in a dataset of $6,931$
patients who originally came from $102$ regions and were diagnosed
with totally $3,178$ unique codes. The inclusion of age and gender
into the model is obvious: they are not only related to and contributing
to the diabetes types, they are also largely associated with other
complications. Information about the regions-of-origin is also important
for diabetes because it is strongly related to the social conditions
and lifestyles, which are of critical importance to the proactive
control of the blood sugar level, which is by far the most cost-effective
method to mitigate diabetes-related consequences.

\paragraph{\noindent Implementation.\label{par:Implementation}}

\noindent Continuous variables are first normalised across the cohort
so that the Gaussian inputs have zero-means and unit variances. We
employ $1$-step contrastive divergence (CD) \cite{hinton_neucom02_training}
for learning. Learning rates vary from type to type and they are chosen
so that reconstruction errors at each data sweep are gradually reduced.
Parameters are updated after each mini-batch of $100$ patients, and
learning is terminated after $100$ data sweeps. The number of hidden
units is determined empirically to be $200$ since large size does
not necessarily improve the clustering/prediction performance.

\noindent For \emph{patient clustering}, once the model has been learned,
the hidden posteriors that are computed using Eq.~(\ref{eq:RBM_factorHid})
can be used as the new representation of the data. To enable fast
bitwise implementation (e.g., see \cite{salakhutdinov_hinton_SIGIR07_semantic}),
we then convert the continuous posteriors into binary activation as
follows: $\hat{h}_{k}=1$ if $P(h_{k}^{1}\mid\bv)\ge\rho_{1}$ and
$\hat{h}_{k}=0$ otherwise for all $k=1,2..,K$ and some threshold
$\rho_{1}\in(0,1)$. We then apply well-known clustering methods including
affinity propagation (AP) \cite{frey_dueck_sci07_ap}, $k$-means
and Bayesian mixture models (BMM). The AP is of particular interest
for our exploratory analysis because it is capable of automatically
determining the number of clusters. It requires the similarity measure
between two patients, and in our binary profiles, a natural measure
is the Jaccard coefficient

\noindent 
\begin{equation}
J\left(p,q\right)=\frac{\mid S\left\{ p\right\} \cap S\left\{ q\right\} \mid}{\mid S\left\{ p\right\} \cup S\left\{ q\right\} \mid}\label{eq:MDA_Clustering_Jaccard}
\end{equation}
where $S\{p\}$ is the set of activated hidden units for patient $p$.
Another hyperparameter is the so-called `preference' which we empirically
set to the average of all pairwise similarities multiplied by $-20$.
This setting gives a reasonable clustering.

\noindent The other two clustering methods require a prior number
of clusters, and here we use the output from the AP. For the $k$-means,
we use the the Hamming distance between activation vectors of the
two patients\footnote{The centroid of each cluster is chosen according to the median elementwise.}.
The BMM is a Bayesian model with multinomial emission probability.

\noindent The task of \emph{disease prediction} is translated into
predicting diagnoses in the future for each patient. We split data
into 2 subsets: The earlier subset, which contains those diagnoses
in the period of $2007-2010$, is used to train the MV.RBM; and the
later subset is used to evaluate the prediction performance. In the
MV.RBM, we order the future diseases according to the probability
that the disease occurs as in Eq.~(\ref{eq:MDA_diseasePrediction}).

\paragraph{Patient Clustering.}

\begin{figure}[h]
\begin{centering}
\includegraphics[scale=0.38]{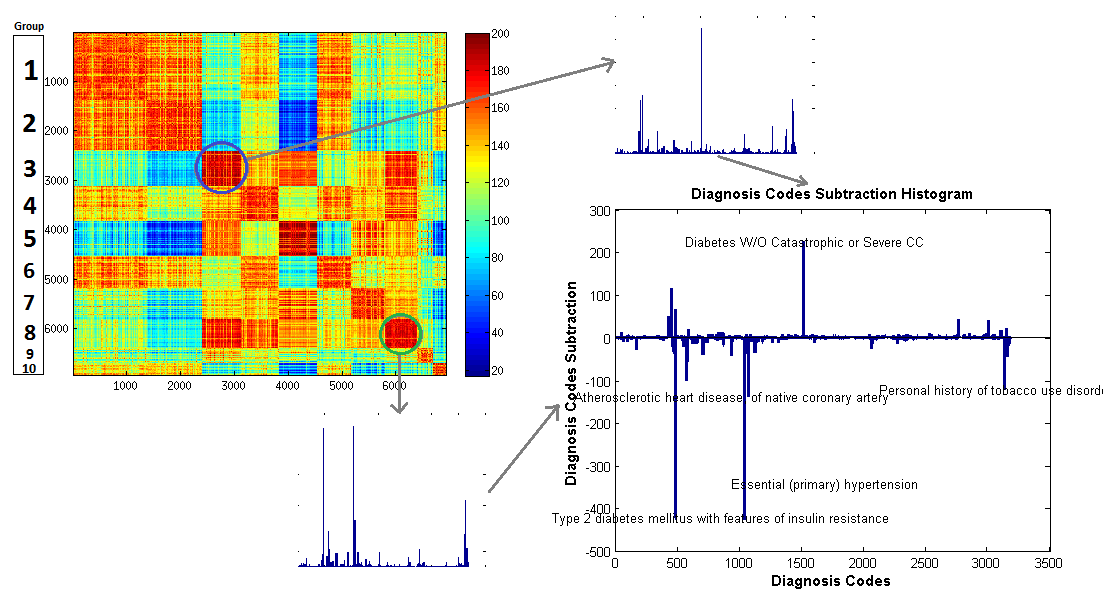}
\par\end{centering}

\caption{Similarity matrix and diagnosis codes histograms. The matrix represents
resemblances of pairwise patients while histograms show quantity of
diagnoses. Group 3 and Group 8 look highly overlapping at the diagnosis
level (top-left figure), but in fact, their clinical conditions are
significantly different when we subtract the two histograms (lower-right
figure). (Best viewed in colors).\label{fig:MDA_SM-DCHist}}
\end{figure}

\begin{figure}[h]
\begin{centering}
\subfloat[Tag cloud of diagnosis descriptions.\label{fig:Type-1-Cloud-Tag}]{\begin{centering}
\includegraphics[width=0.75\textwidth]{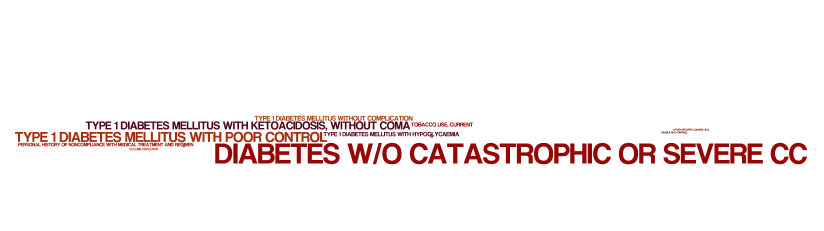}
\par\end{centering}

}\subfloat[Age histogram.\label{fig:MDA_Type-1Age-histogram}]{\begin{centering}
\includegraphics[width=0.2\textwidth]{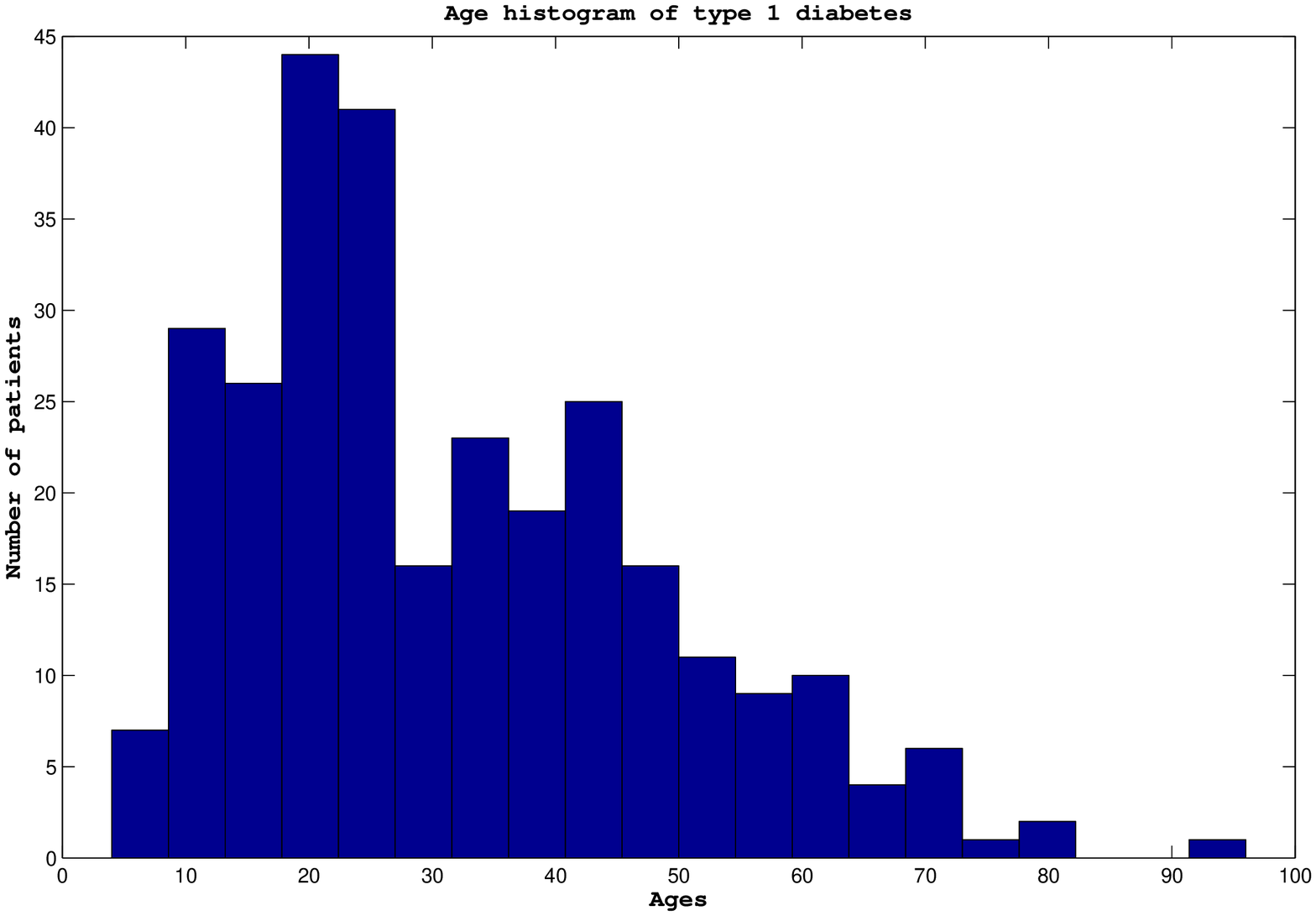}
\par\end{centering}

}
\par\end{centering}

\caption{Type I diabetes mellitus: Primary diagnoses and age distribution.
Two figures confirms the existing knowledge that Type I diabetes mellitus
often occurs in the younger population.\label{fig:MDA_Type-1-diabetes}}
\end{figure}

\begin{figure}[h]
\begin{centering}
\subfloat[Tag cloud of diagnosis descriptions.\label{fig:MDA_Type-2-Cloud-Tag}]{\begin{centering}
\includegraphics[width=0.75\textwidth]{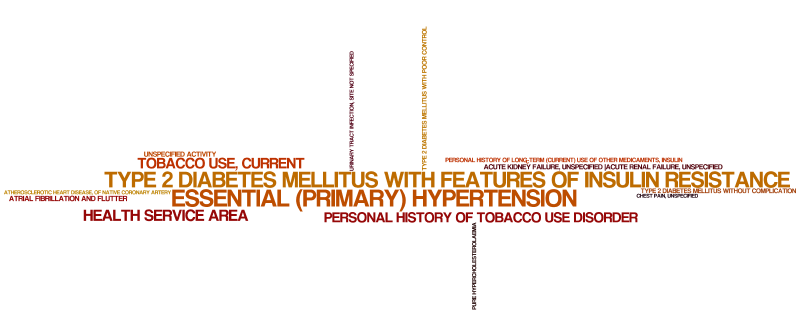}
\par\end{centering}

\centering{}}\subfloat[Age histogram.\label{fig:MDA_Type-2-Age-histogram}]{\begin{centering}
\includegraphics[width=0.2\textwidth]{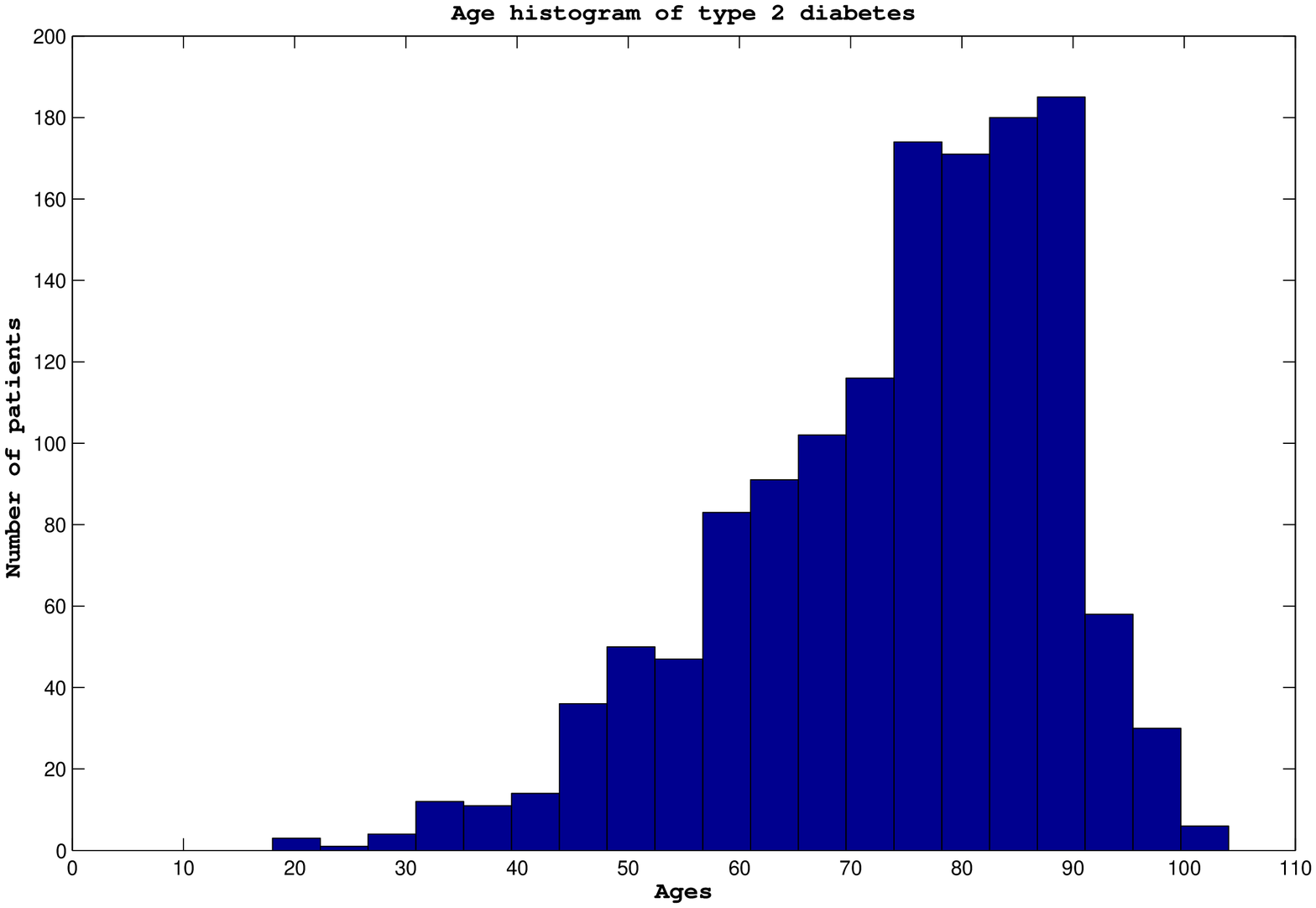}
\par\end{centering}

}
\par\end{centering}

\caption{Type II Diabetes mellitus: Primary diagnoses and age distribution.
We can see that the age distribution is distinct from the Type I group.\label{fig:MDA_Type-2-diabetes}}
\end{figure}

\begin{figure}[h]
\begin{centering}
\subfloat[\label{fig:MDA_t-SNE}]{\begin{centering}
\includegraphics[width=0.48\textwidth]{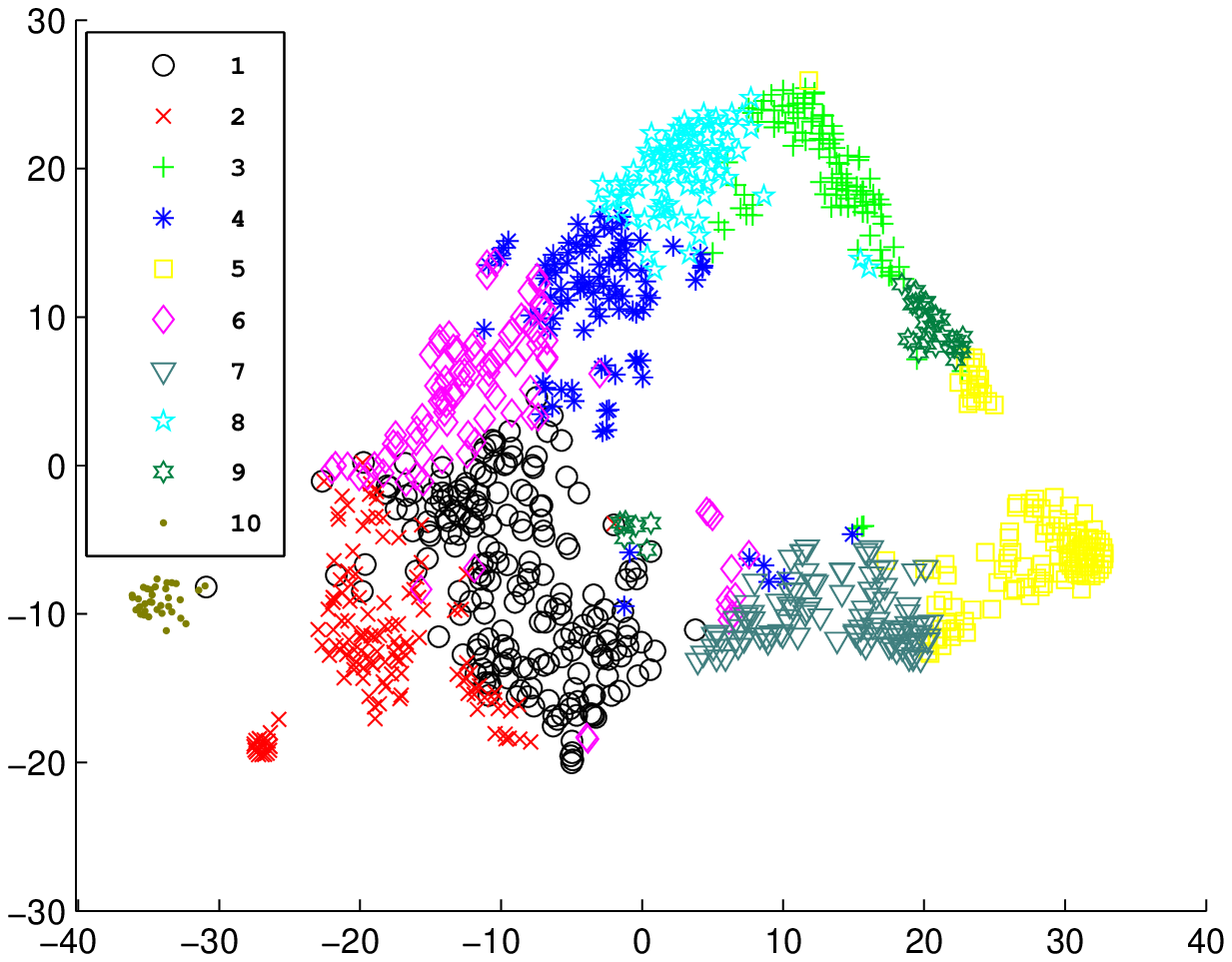}
\par\end{centering}

}\subfloat[\label{fig:MDA_RandIndex}]{\centering{}\includegraphics[width=0.48\textwidth]{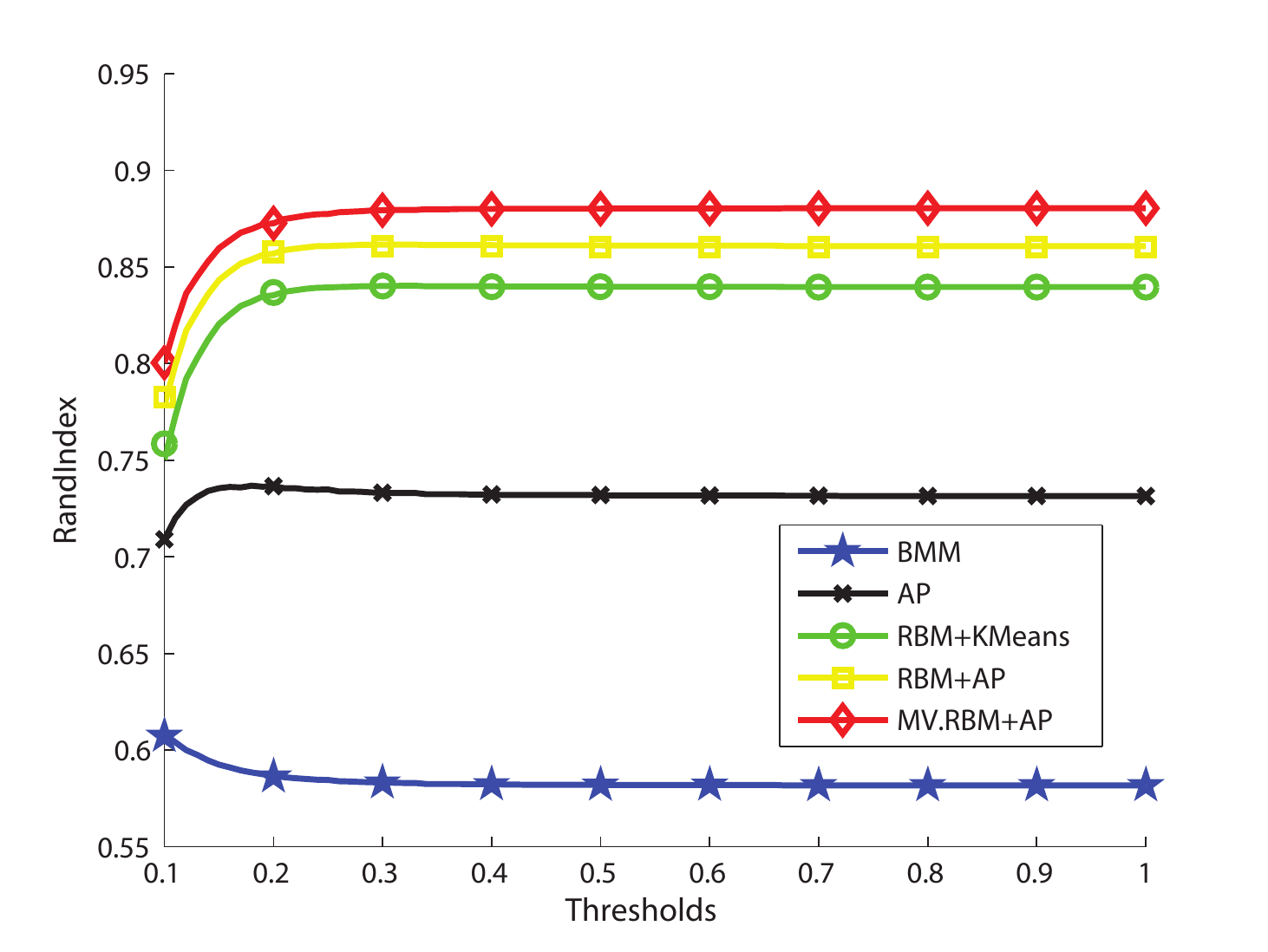}}
\par\end{centering}

\caption{Visualisation and quantitative assessment of clusters. (a) t-SNE \cite{van_hinton_jmlr08_tsne}
projection on $2,000$ latent profiles. Groups are labelled by the
outputs of the AP. (Best viewed in colors). (b) Rand-index curves
in patient clustering. \emph{AP}: affinity propagation, \emph{BMM}:
Bayesian mixture model, \emph{RBM}: MV.RBM with diagnosis codes only. }
\end{figure}

\noindent First we wish to validate that the latent profiles discovered
by the MV.RBM are informative enough so that \emph{clinically meaningful}
clusters can be formed. Figure~\ref{fig:MDA_SM-DCHist} shows the
$10$ clusters returned by the AP and the similarity between every
patient pair (depicted in colour, where the similarity increases with
from blue to red). It is interesting that we are able to discover
a group whose conditions are mostly related to Type I diabetes (Figures~\ref{fig:Type-1-Cloud-Tag}
and \ref{fig:MDA_Type-1Age-histogram}), and another group associated
with Type II diabetes (Figures~\ref{fig:MDA_Type-2-Cloud-Tag} and
\ref{fig:MDA_Type-2-Age-histogram}). The grouping properties can
be examined visually using a visualisation tool known as t-SNE \cite{van_hinton_jmlr08_tsne}
to project the latent profiles onto 2D. Figure~\ref{fig:MDA_t-SNE}
depicts the distribution of patients, where the colours are based
on the group indices assigned earlier by the AP.

For quantitative evaluation, we calculate the \emph{Rand-index} \cite{rand_jasa71_objective}
to assess the quality of resulting clusters, given that we do not
have cluster labels. The Rand-index is the pairwise accuracy between
any two patients. To judge whether two patients share the same cluster,
we consult the diagnosis code hierarchy of the ICD-10 \cite{WHO_ICD-10}\textcolor{black}{.
We use hierarchical assessment since a diagnosis code may have multiple
levels. E11.12, for example, has two levels: E11 and E11.12. The lower
level code specifies disease more clearly whilst the higher is more
abstract. Therefore we have two ways for pairwise judgement: the Jaccard
coefficient (Par.~\ref{par:Implementation}) and code }`\textcolor{black}{cluster'
which is the grouping of codes that belong to the same disease class,
as specified by the latest WHO standard ICD-10. At the lowest level,
two patients are considered similar if the two corresponding code
sets are sufficiently overlapping, i.e.,} their\textcolor{red}{{} }Jaccard
coefficient is greater than a threshold $\rho_{2}\in(0,1)$. At higher
level, on the other hand, we consider two patients to be clinically
similar if they share higher level diabetes code of the same code
`\textcolor{black}{cluster}'. For instance, two patients with two
codes E11.12 and E11.20 are similar at the level E11\footnote{This code group is for non-insulin-dependent \emph{diabetes mellitus}.},
but they are dissimilar at the lower level. Note that this hierarchical
division is for evaluation only. We use codes at the lowest level
as replicated softmax units in our model (Section~\ref{sub:MDA_Framework}).

Figure~\ref{fig:MDA_RandIndex} reports the Rank-indices with respect
to the assessment at the lowest level in the ICD-10 hierarchy for
clustering methods with and without MV.RBM pre-processing. \textcolor{black}{At
the next ICD-10 level, the MV.RBM/AP achieves a Rand-index of $0.6040$,
which is, again, the highest among all methods, e.g., using the RBM/AP
yields the score of $0.5870$, and using AP on diagnosis codes yields
$0.5529$. }This clearly demonstrates that (i) \textcolor{black}{MV.RBM}
latent profiles would lead to better clustering that those using diagnosis
codes directly, and (ii) modelling mixed-types would be better than
using just one input type (e.g., the diagnosis codes).

\paragraph{Disease Prediction.}

\begin{figure}
\begin{centering}
\includegraphics[width=0.8\textwidth]{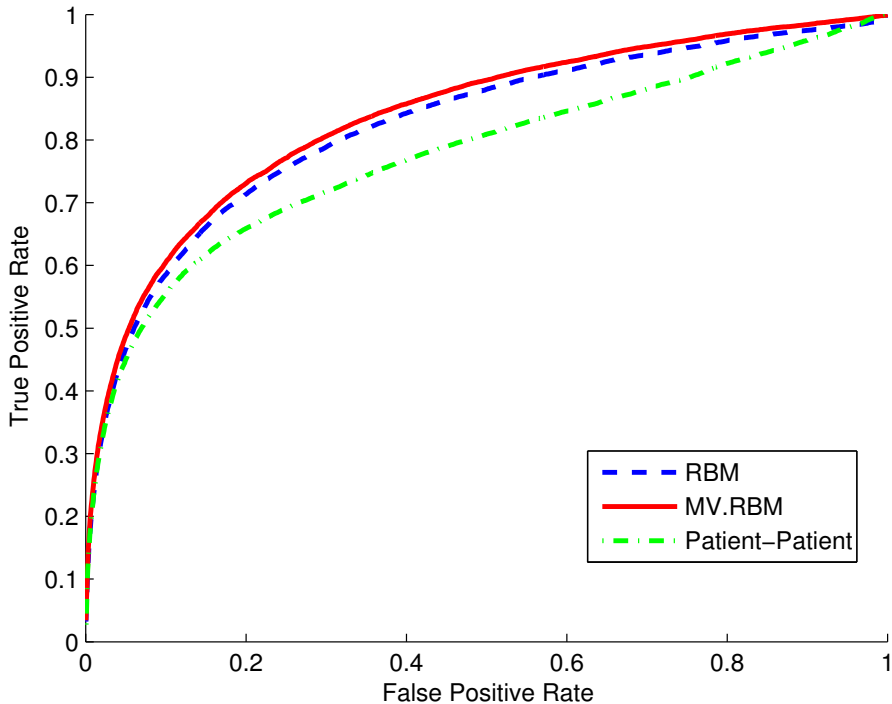}
\par\end{centering}

\caption{ROC curves in disease prediction. \emph{RBM} is MV.RBM with diagnosis
codes only; \emph{Patient-Patient} is the $k$-nearest neighbours
method. Best viewed in colors.\label{fig:MDA_Prediction-ROC-Curve}}
\end{figure}

The prediction results are summarised in Figure~\ref{fig:MDA_Prediction-ROC-Curve},
where the ROC curve of the MV.RBM is compared against that of the
baseline using $k$-nearest neighbours ($k$-NN). The $k$-NN approach
to disease ranking is as follows: For each patient, a neighbourhood
of the $50$ most similar patients is collected based on the Jaccard
coefficient over sets of unique diagnoses. The diagnoses are then
ranked according to their occurrence frequency within the neighbourhood.
As can be seen from the figure, the latent profile approaches outperform
the $k$-NN method. The MV.RBM with contextual information such as
age, gender and region-of-birth proves to be useful. In particular
the the areas under the ROC curve (AUC) of the MV.RBMs are $0.84$
(with contextual information) and $0.82$ (without contextual information).
These are clearly better than the score obtained by $k$-NN, which
is $0.77$.

\subsection{Image Retrieval\label{sub:IR}}

Images are typically retrieved based on the distance from the query
image. Thus the retrieval quality depends critically on how the images
are represented and the distance metric operating on the representation
space. Standard vector-based representations may involve colour histograms
or visual descriptors; and distance metrics can be those working in
the vector space. However, they suffer from important drawbacks. First,
there are no simple ways to integrate multiple representations (e.g.,
histograms and bag-of-words) and multiple modalities (e.g., visual
words and textual tags). Furthermore, designing a representation separately
from distance metric is sup-optimal -- it takes time to search for
the best distance metric for a given representation. And third, using
low-level features may not capture well the high-level semantics of
interest, leading to poor retrieval quality if the visual features
are similar and the objects are different. For example, it can be
easy to confuse between a lion and a wolf if we rely on the textures
and colours alone.

In this problem, our solution is to \emph{learn} both the higher representation
and the distance metric specifically for the retrieval task. The higher
representation would capture the regularities and factors of variation
in the data space from multiple lower feature types and modalities.
At the same time, the representation would lead to small distances
between conceptually related objects and large distances between those
unrelated. To that end, we integrate structured sparsity (Section~\ref{sub:Sparsity})
and distance metric learning (Section~\ref{sub:metric_learning})
into MV.RBM (Section~\ref{sub:MV.RBM}). The MV.RBM is a probabilistic
architecture capable of integrating several data types into a homogeneous
``latent'' representation in an unsupervised fashion. Our methods
include the introduction of the counting of visual/textual words \cite{salakhutdinov_hinton_SIGIR07_semantic}
and the group-wise sparsity \cite{luo_etal_aaai11_sgrbm} into the
model. During the training phase, the model learning is regularised
in the way that the information theoretic distances on the latent
representation between intra-concept images are minimised and those
between inter-concept images are maximised. During the testing phase,
the learned distance metrics are then used for retrieving similar
images.

We demonstrate the effectiveness of the proposed method on the NUS-WIDE
data. This data is particularly rich: Each image has multiple visual
representations, sometimes social tags, and one or more manually annotated
high-level concepts. We run experiments on two subsets. The first
is the well-studied $13$ animal subset in which we show that our
method is competitive against recent state-of-the-arts. The second
subset contains $20,000$ single-concept images. We obtain $79.5\%$
improvement on mean average precision (MAP) over the standard nearest
neighbours approach and $45.7\%$ increase on the normalised discounted
cumulative gain (NDCG).

Section~\ref{sub:IR_IntgFramework} presents the Mixed-Variate RBM
with group-wise sparsity and metric learning for image retrieval.
We show empirical evaluation on the two NUS-WIDE subsets in Section~\ref{sub:IR_Exp}.

\subsubsection{Integrated framework\label{sub:IR_IntgFramework}}

We present our framework of simultaneous learning of sparse data representation
and distance metric for image retrieval tasks. The framework has three
components: (i) a mixed-variate machine that maps multiple feature
types and modalities into a homogeneous higher representation, (ii)
a regulariser that promotes structured sparsity on the learned representation,
and (iii) an information-theoretic distance operating on the learned
representation. To what follows, we involve the structure sparsity
in Section~\ref{sub:Sparsity} and metric learning in Section~\ref{sub:metric_learning}
to the unsupervised learning parameters of MV.RBM in Section~\ref{sub:MV.RBM}.
Our approach has three goals: Capturing the joint representation of
visual and textual features by maximising the data likelihood $\likelihood\left(\bv;\psi\right)=\sum_{\bh}P\left(\bh,\bv;\psi\right)$,
enhancing structural sparsity through minimising $\mathcal{R}\left(\bv\right)$,
and regularising intra-concept and inter-concept distance metrics
$\dataset_{\mathsf{N}\left(f\right)}$ and $\dataset_{\bar{\mathsf{N}}\left(f\right)}$.
The objective function is now the following regularised likelihood

\[
\likelihood_{reg}=\sum_{f}\log\likelihood\left(\bv^{\left(f\right)};\psi\right)-\alpha\mathcal{R}\left(\bv\right)-\beta\left(\sum_{f}\dataset_{\mathsf{N}\left(f\right)}-\sum_{f}\dataset_{\bar{\mathsf{N}}\left(f\right)}\right)
\]

\noindent where $\alpha\geq0$ is the regularising constant for sparsity,
$\beta\geq0$ is the coefficient to control the effect of distance
metrics. Maximising this regularised likelihood is equivalent to simultaneously
maximising the data likelihood $\likelihood\left(\bv;\psi\right)$,
minimising the regularisation function $\mathcal{R}\left(\bv\right)$,
minimising the neighbourhood distance $\dataset_{\mathsf{N}\left(f\right)}$
and maximising the non-neighbourhood distance $\dataset_{\bar{\mathsf{N}}\left(f\right)}$.

In Section~\ref{sec:aims_approach}, the derivatives of regularised
likelihood with respect to parameters are all specified. Consequently,
we perform stochastic gradient ascent to update parameters as follows

\[
\psi\leftarrow\psi+\lambda\left(\frac{\partial}{\partial\psi}\mathcal{L}_{reg}\right)
\]

\noindent for some learning rate $\lambda>0$.

\subsubsection{Empirical evaluations\label{sub:IR_Exp}}

In this task, we quantitatively evaluate our method on two real datasets.
Both datasets are subsets selected from the NUS-WIDE dataset \cite{chua_etal_civr09_nuswide},
which was collected from Flickr. The NUS-WIDE dataset includes $269,648$
images which are associated with $5,018$ unique tags. There are $81$
concepts in total. For each image, six types of low-level features
\cite{chua_etal_civr09_nuswide} are extracted, including $64$-D
color histogram in LAB color space, $144$-D color correlogram in
HSV color space, $73$-D edge direction histogram, $128$-D wavelet
texture, $225$-D block-wise LAB-based color moments extracted over
$5\times5$ fixed grid partitions and $500$-D bag-of-word (BOW) based
on SIFT descriptions.

For training our model, mapping parameters $\bW$ are randomly initialised
from small normally distributed numbers, i.e. Gaussian $\mathcal{N}\left(0;0.01\right)$,
and biases $\left(\ba,\bb\right)$ are set to zeros. To enhance the
speed of training, we divide training images into small ``mini-batches''
of $B=100$ images. Hidden and visible learning rates are fixed to
$0.02$ and $0.3$, respectively\footnote{Learning rate settings are different since the hidden are binary whilst
the visible are unbounded.}. Parameters are updated after every mini-batch and the learning finishes
after $100$ scans through the whole data. Once parameters have been
learned, images are projected onto the latent space using Eq.~(\ref{eq:RBM_factorHid}).
We set the number of hidden groups $M$ to the expected number of
groups (e.g., concepts), and the number of hidden units $K$ is multiple
of $M$. The retrieved images are ranked based on the negative KL-div
on these latent representations. We repeat $10$ times and report
the mean and standard deviation of the performance measures.

\paragraph{Retrieving Animals.}

\begin{figure}[t]
\begin{minipage}[t]{0.45\textwidth}%
\noindent \begin{center}
\begin{tabular}{|ccccccc|}
\multicolumn{1}{c}{} &  &  &  &  &  & \multicolumn{1}{c}{}\tabularnewline
\hline 
\hline 
\textbf{\scriptsize{}squirrel} & \textbf{\scriptsize{}cow} & \textbf{\scriptsize{}cat} & \textbf{\scriptsize{}tiger} & \textbf{\scriptsize{}whale} & \textbf{\scriptsize{}rabbit} & \textbf{\scriptsize{}snake}\tabularnewline
\includegraphics[scale=0.2]{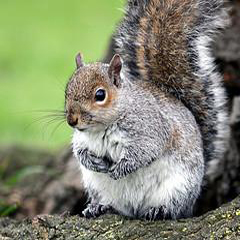} & \includegraphics[scale=0.2]{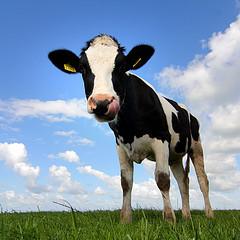} & \includegraphics[scale=0.2]{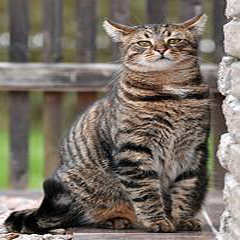} & \multirow{3}{*}{\includegraphics[scale=0.2]{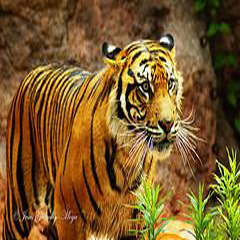}} & \includegraphics[scale=0.2]{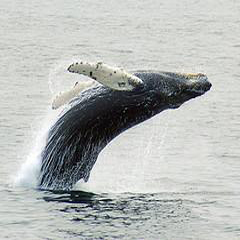} & \includegraphics[scale=0.2]{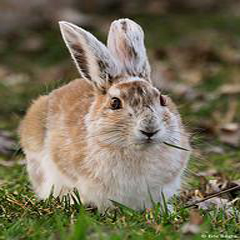} & \includegraphics[scale=0.2]{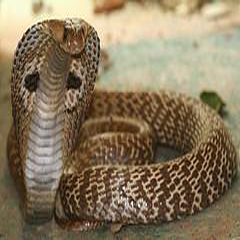}\tabularnewline
\textbf{\scriptsize{}zebra} & \textbf{\scriptsize{}lion} & \textbf{\scriptsize{}elephant} &  & \textbf{\scriptsize{}antler} & \textbf{\scriptsize{}wolf} & \textbf{\scriptsize{}hawk}\tabularnewline
\includegraphics[scale=0.2]{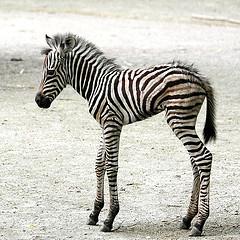} & \includegraphics[scale=0.2]{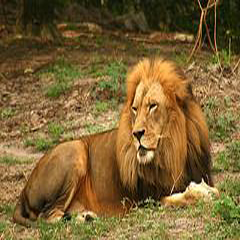} & \includegraphics[scale=0.2]{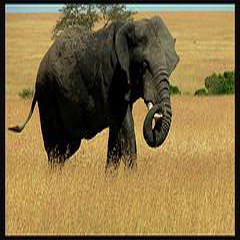} &  & \includegraphics[scale=0.2]{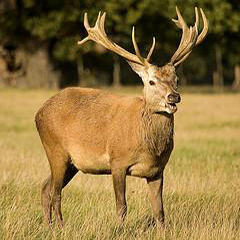} & \includegraphics[scale=0.2]{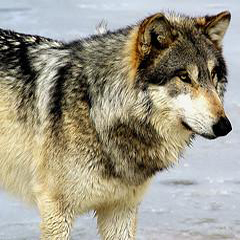} & \includegraphics[scale=0.2]{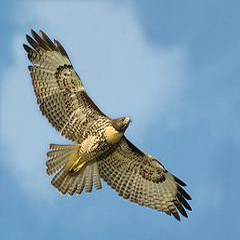}\tabularnewline
\hline 
\hline 
\multicolumn{1}{c}{} &  &  &  &  &  & \multicolumn{1}{c}{}\tabularnewline
\end{tabular}
\par\end{center}%
\end{minipage}\hfill{}%
\begin{minipage}[t]{1\columnwidth}%
\noindent \begin{center}
\vspace{0.016\textheight}
\includegraphics[scale=0.3]{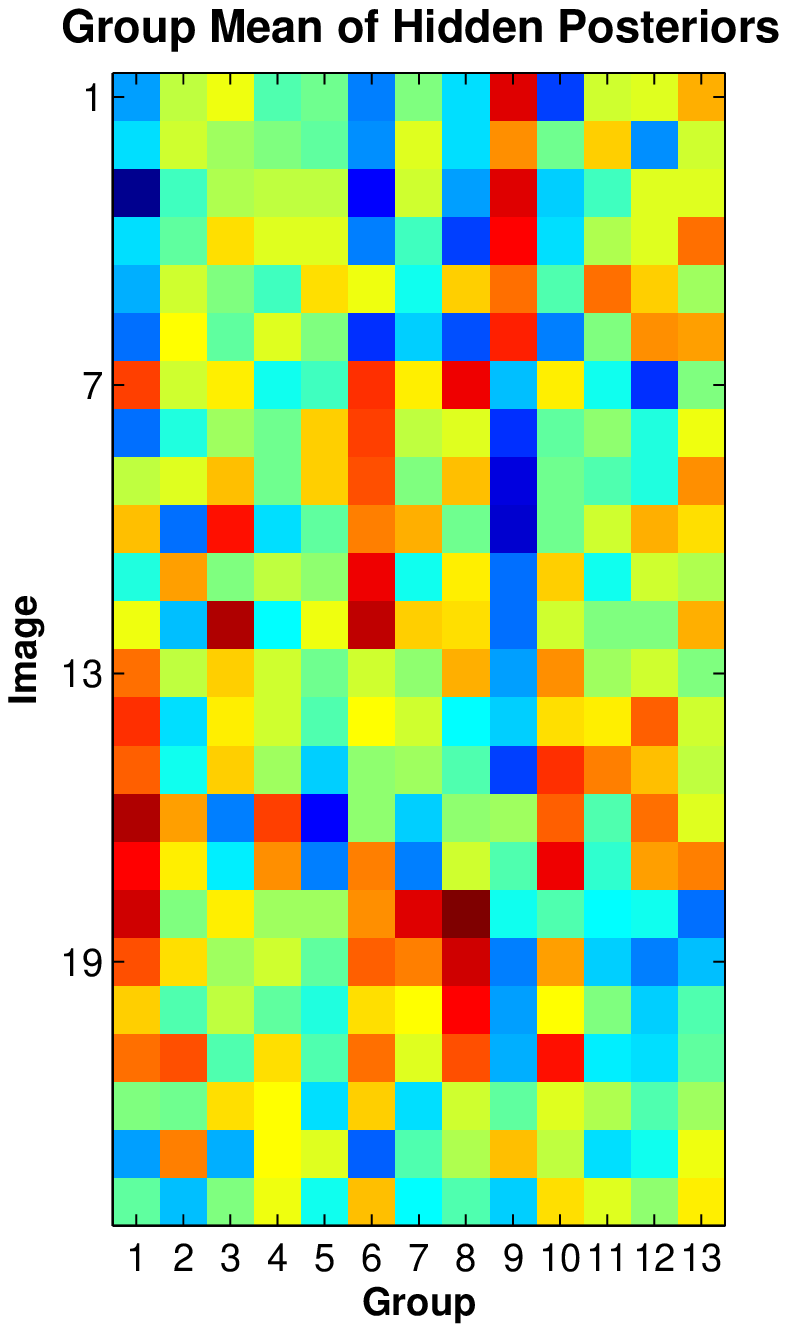}
\par\end{center}%
\end{minipage}

\vspace{-0.02\textheight}
\caption{Example images of each species in NUS-WIDE animal dataset. The last
column shows the group mean of hidden posteriors by colours, one line
per image. The red cells illustrate higher values whilst the blue
denote the lower. It is clear that 4 groups of 6 consecutive images
form 4 strips in different groups (9,6,1,8).\label{fig:IR_EgImg_NUS-WIDE-animal.}}
\end{figure}

The first subset is the NUS-WIDE animal dataset which contains $3,411$
images of $13$ animals - \emph{squirrel, cow, cat, zebra, tiger,
lion, elephant, whale, rabbit, snake, antler, wolf and hawk}. Figure~\ref{fig:IR_EgImg_NUS-WIDE-animal.}
shows example images of each category. Out of $3,411$ images, $2,054$
images are used for training and the remaining for testing. In the
testing phase, each test image is used to query images in training
set to receive a list of images ranked basing on similarities. These
settings are identical to those used in previous work \cite{gupta2011bayesian,gupta2012slice,Chen2010}.
For our methods, the similarity measure is negative symmetric Kullback-Leibler
divergence (KL-div) learned from data (Section~\ref{sub:IR_IntgFramework}).
The retrieval performance is evaluated using Mean Average Precision
(MAP) over all received images in training set. Two images are considered
similar if they depict the same type of animal.

In this experiment, we concatenate first five histogram features of
each animal image into a long vector and ignore BOW features to fairly
compare with recent work. Thus we treat elements of the vector as
Gaussian units and normalise them across all training images to obtain
zeros mean and unit standard variance. Note that the MV.RBM here reduces
to the plain RBM with single Gaussian type.
\begin{figure}[h]
\begin{centering}
\subfloat{\centering{}\includegraphics[width=0.33\textwidth]{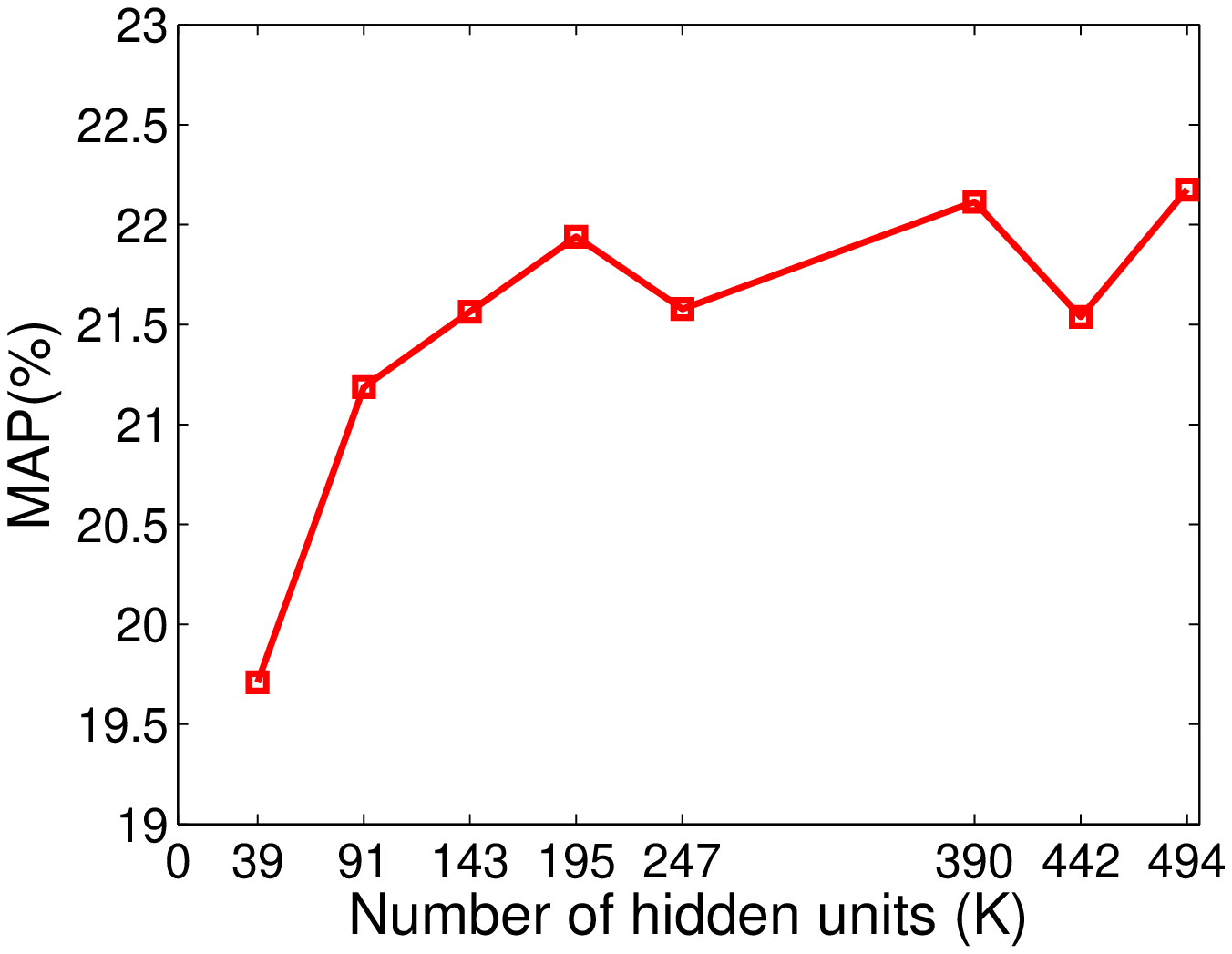}}\subfloat{\centering{}\includegraphics[width=0.33\textwidth]{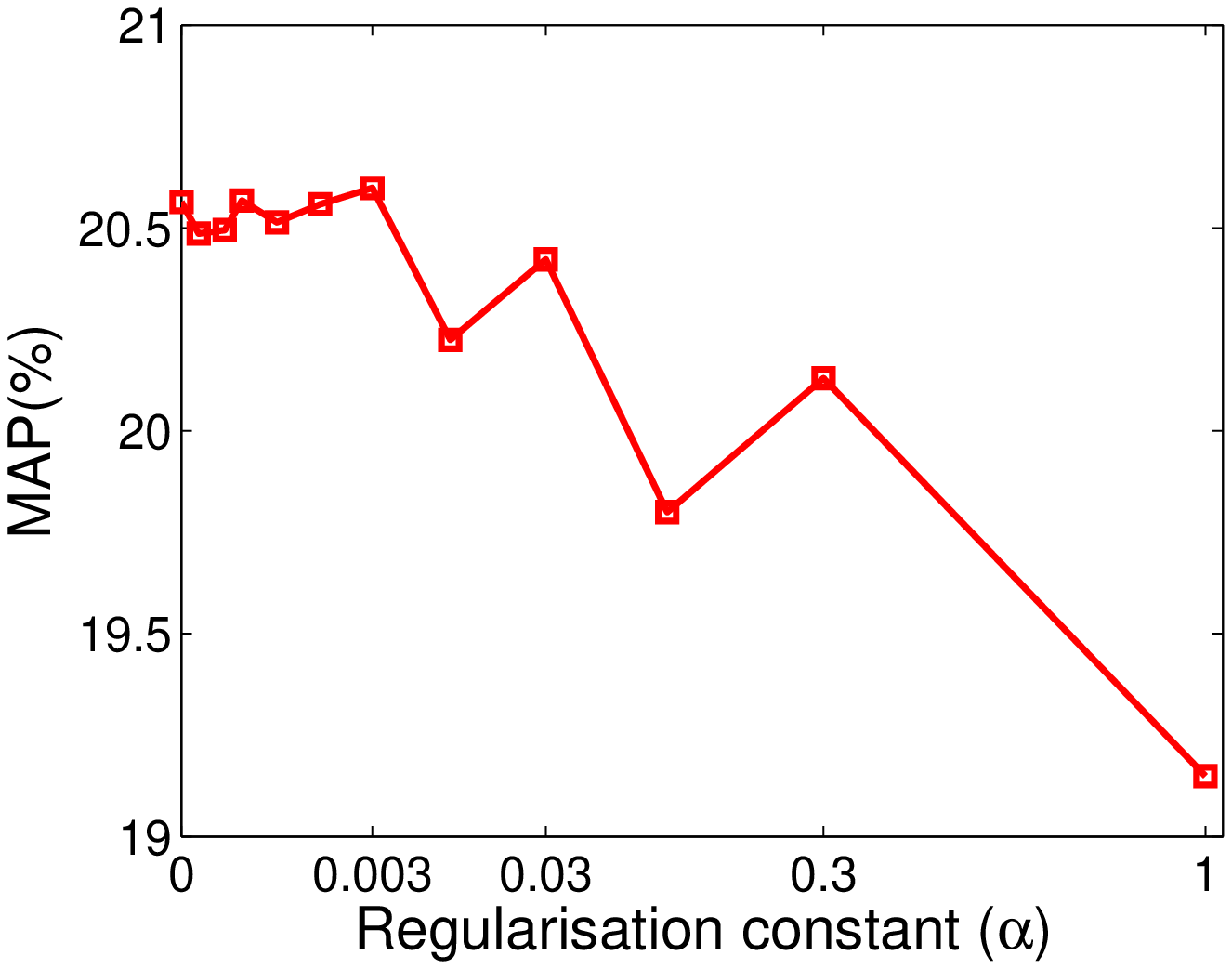}}\subfloat{\centering{}\includegraphics[width=0.33\textwidth]{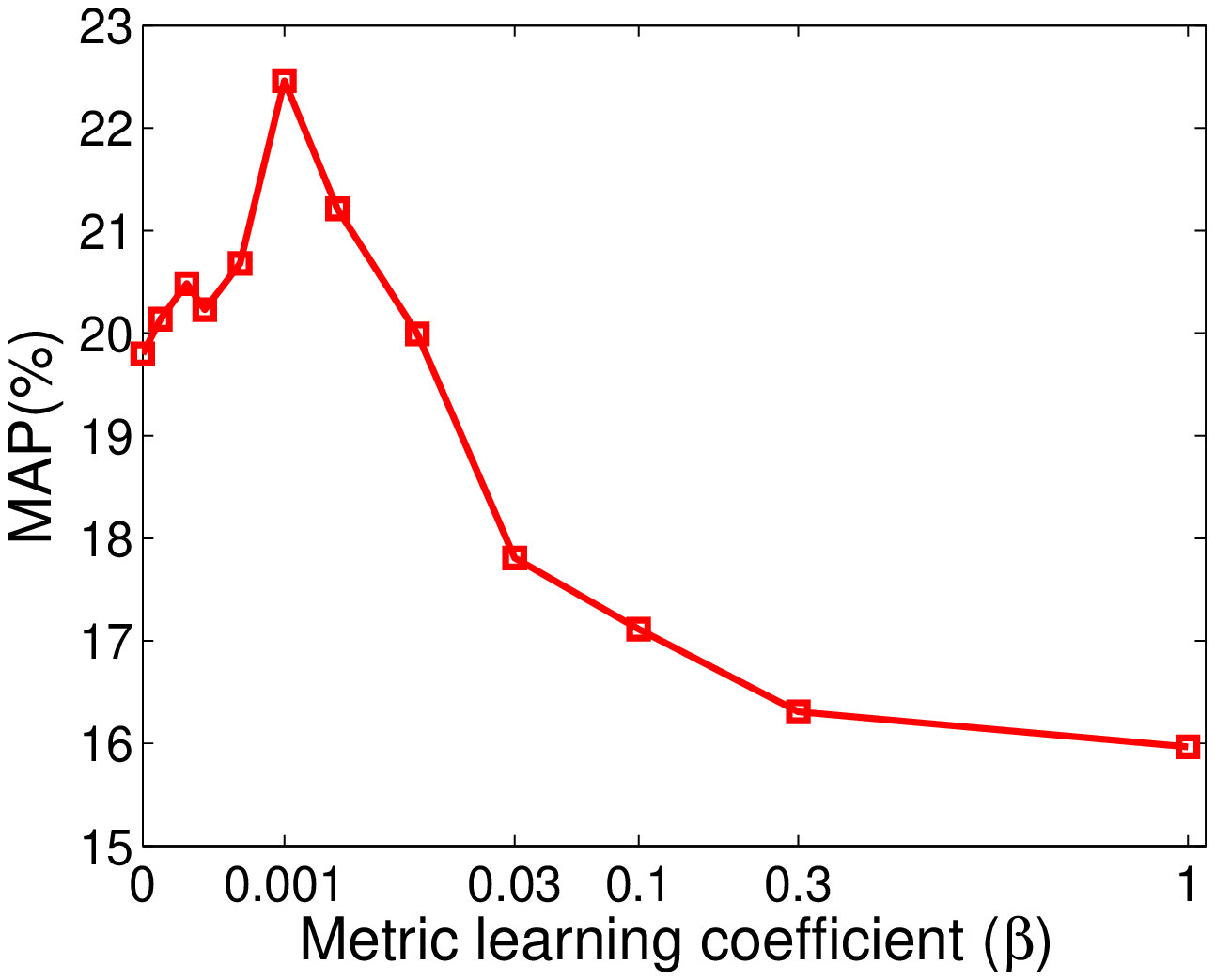}}
\par\end{centering}

\caption{The MAP performance (\%) as functions of the hyper-parameters: (Left)
The number of hidden units $K$ (with $\alpha=0.003$ and $\beta=0.001$);
(Middle) The regularisation constant $\alpha$ (with $K=195$); (Right)
The metric learning coefficient $\beta$ (with $K=195$ and $\alpha=0.003$).\label{fig:IR_MAP_numHUs_NormRate_MetW}}
\end{figure}

To find the best setting of the hyper-parameters $\alpha,\beta$ and
$K$, we perform initial experiments with varying values. Figure~\ref{fig:IR_MAP_numHUs_NormRate_MetW}
reports the MAP performance (\%) with respect to these values. Here
$\alpha=0$ means no sparsity constraint and $\beta=0$ means no metric
learning. As can be seen from the left figure, the performance stops
increasing after some certain hidden size. Adding certain amount of
sparsity control slightly improves the result (see the middle figure).
A much stronger effect is due to metric learning, as shown in the
right figure. From these observations, we choose $K=195$ (15 units
per group), $\alpha=0.003$ and $\beta=0.001$.
\begin{figure}[H]
\begin{centering}
\begin{tabular}{ccccc}
 &  &  &  & \tabularnewline
\hline 
\hline 
 &  & \textbf{\small{}$k$-NN} &  & \tabularnewline
\textbf{\textcolor{black}{\emph{\footnotesize{}lion}}} & \textcolor{red}{\footnotesize{}wolf} & \textcolor{blue}{\footnotesize{}lion} & \textcolor{red}{\footnotesize{}wolf} & \textcolor{blue}{\footnotesize{}lion}\tabularnewline
\includegraphics[scale=0.33]{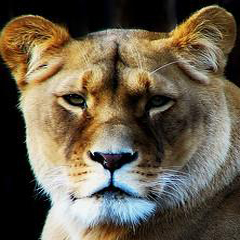} & \includegraphics[scale=0.33]{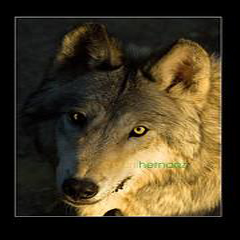} & \includegraphics[scale=0.33]{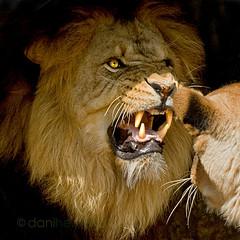} & \includegraphics[scale=0.33]{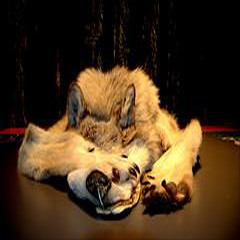} & \includegraphics[scale=0.33]{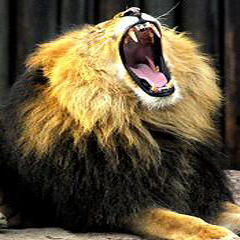}\tabularnewline
\hline 
\multicolumn{5}{c}{\textbf{\small{}RBM}}\tabularnewline
\textbf{\textcolor{black}{\emph{\footnotesize{}lion}}} & \textcolor{red}{\footnotesize{}wolf} & \textcolor{blue}{\footnotesize{}lion} & \textcolor{blue}{\footnotesize{}lion} & \textcolor{red}{\footnotesize{}wolf}\tabularnewline
\includegraphics[scale=0.33]{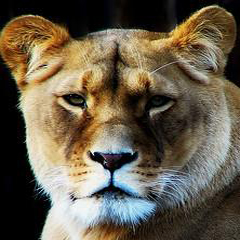} & \includegraphics[scale=0.33]{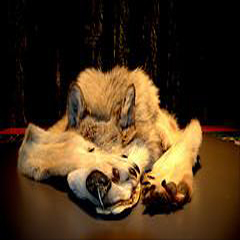} & \includegraphics[scale=0.33]{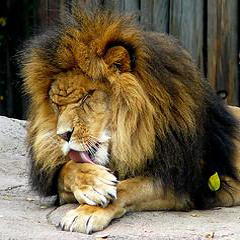} & \includegraphics[scale=0.33]{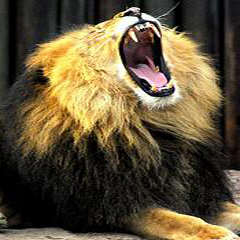} & \includegraphics[scale=0.33]{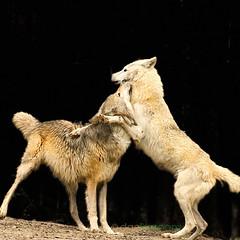}\tabularnewline
\hline 
\multicolumn{5}{c}{\textbf{\small{}RBM+SG}}\tabularnewline
\textbf{\textcolor{black}{\emph{\footnotesize{}lion}}} & \textcolor{blue}{\footnotesize{}lion} & \textcolor{red}{\footnotesize{}wolf} & \textcolor{blue}{\footnotesize{}lion} & \textcolor{blue}{\footnotesize{}lion}\tabularnewline
\includegraphics[scale=0.33]{figs/retrieval/rbm/609} & \includegraphics[scale=0.33]{figs/retrieval/rbm/777} & \includegraphics[scale=0.33]{figs/retrieval/rbm/1781} & \includegraphics[scale=0.33]{figs/retrieval/rbm/865} & \includegraphics[scale=0.33]{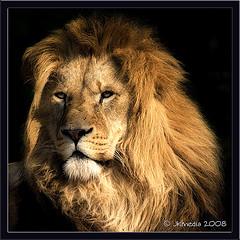}\tabularnewline
\hline 
\multicolumn{5}{c}{\textbf{\small{}RBM+SG+ML}}\tabularnewline
\textbf{\textcolor{black}{\emph{\footnotesize{}lion}}} & \textcolor{blue}{\footnotesize{}lion} & \textcolor{blue}{\footnotesize{}lion} & \textcolor{blue}{\footnotesize{}lion} & \textcolor{blue}{\footnotesize{}lion}\tabularnewline
\includegraphics[scale=0.33]{figs/retrieval/rbm/609} & \includegraphics[scale=0.33]{figs/retrieval/rbm/777} & \includegraphics[scale=0.33]{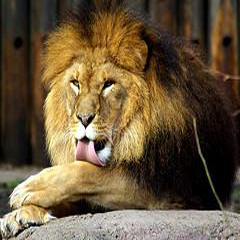} & \includegraphics[scale=0.33]{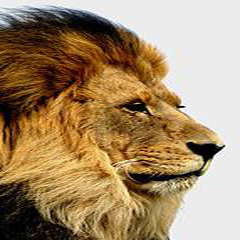} & \includegraphics[scale=0.33]{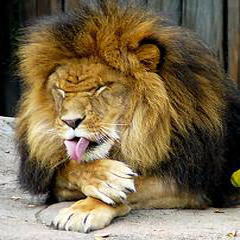}\tabularnewline
\hline 
\hline 
 &  &  &  & \tabularnewline
\end{tabular}
\par\end{centering}

\caption{Retrieved images for query image of a lion in testing set. $k$-NN:
$k$-nearest neighbours, RBM+SG+ML: RBM with sparse group, metric
learning. First column is the queried images. Blue titles are right
retrieval whilst the red are wrong. Four retrieved images are sorted
in descent order of similarities from left to right.\label{fig:IR_Retrieval_Results}}
\end{figure}

Figure~\ref{fig:IR_Retrieval_Results} shows how structured sparsity
and metric learning contributes to the higher retrieval quality. The
naive nearest neighbour on concatenated normalised features\footnote{Each feature is normalised to zero mean and unit standard variance
over images.} would confuse a wolf with the query of lion, possibly due to the
similar colour profiles. The standard RBM admits the same error suggesting
that learning just regularities is not enough. Adding structured sparsity
(RBM+SG) corrects one error and using learned metric (RBM+SG+ML) would
correct all the errors.
\begin{table}
\centering{}\caption{Image retrieval results to compare with recent state-of-the-art multiview
learning and hierarchical modelling techniques on NUS-WIDE animal
dataset. RBM+SG+ML is RBM with latent sparse groups and metric learning.\label{tab:IR_NUS_WIDE_Animal}}
\begin{tabular}{|c|c|}
\hline 
\multirow{1}{*}{\emph{Method}} & \multicolumn{1}{c|}{\emph{MAP}}\tabularnewline
\hline 
\hline 
DWH \cite{Chen2010} & 0.153\tabularnewline
\hline 
TWH \cite{Chen2010} & 0.158\tabularnewline
\hline 
MMH \cite{Chen2010} & 0.163\tabularnewline
\hline 
NHFA-GGM (approx.) \cite{gupta2011bayesian} & 0.179\textpm 0.013\tabularnewline
\hline 
Proposed NHFA-GGM \cite{gupta2012slice} & 0.195\textpm 0.013\tabularnewline
\hline 
\emph{RBM} & \emph{0.199\textpm 0.001}\tabularnewline
\hline 
\hline 
\emph{RBM+SG} & \emph{0.206}\textpm 0.002\tabularnewline
\hline 
\textbf{\emph{RBM+SG+ML}} & \textbf{\emph{0.252}}\textpm \textbf{\emph{0.002}}\tabularnewline
\hline 
\end{tabular}
\end{table}

Finally, Table \ref{tab:IR_NUS_WIDE_Animal} presents the MAP results
of our methods (RBM, RBM with sparse group (SG) and with metric learning
(ML)) in comparison with recent work \cite{Chen2010,gupta2011bayesian,gupta2012slice}
on the NUS-WIDE animal dataset. It is clear that RBM and RBM with
SG are competitive against all previous methods; and RBM integrated
with SG and ML significantly outperforms state-of-the-art approaches.

\paragraph{Retrieving Individual Concepts.}

In the second experiment, we aim to demonstrate the capability of
our method to handle heterogeneous types of features and larger data.
We randomly pick 10,000 images for training and 10,000 for testing.
Each image in this subset has exactly one concept and altogether,
they cover the entire 81 concepts of the NUS-WIDE dataset. Six visual
features (1 bag-of-word and 5 histogram-like) and associated social
tags, limited to 1,000, of each image are taken. The MV.RBM encodes
5 histogram features as Gaussian, social tags as binary and BOW as
Poisson units. We further transform counts into $log$ space using
$\left[log\left(1+\textnormal{count}\right)\right]$. 

Besides MAP score, we also compute the Normalized Discounted Cumulative
Gain (NDCG) \cite{Jarvelin:2002:CGE:582415.582418} for evaluation.
Here we only use the top 100 similar images for calculating MAP and
top 10 images for computing NDCG. We create 2 baselines and 4 versions
of our approach to show the improvement of the MV.RBM when adding
sparse groups and metric learning (MV.RBM+SG+ML). The first baseline
is to employ $k$-NN method on concatenated feature vectors. First
features are normalised to zeros mean and unit vector over images
to eliminate the differences in dimensionality and scale. The second
baseline is fusion of multiple plain RBMs, each of which is type specific,
i.e., BOW as Poisson, visual histograms as Gaussian and textual tags
as binary. For each type of RBM, visible input data is mapped into
binary latent representation. Then these latent representations are
concatenated into a single latent representation.

The first version (RBM+SG) is the second baseline with group-wise
sparsity (Section~\ref{sub:Sparsity}). The second version (MV.RBM)
jointly models all 7 types of features. The third version (MV.RBM+SG)
is MV.RBM with $81$ sparse groups (Section~\ref{sub:Sparsity}).
And finally, the proposed solution (MV.RBM+SG+ML) integrates both
the sparsity and the metric learning into the MV.RBM.
\begin{table}[t]
\caption{Comparison of image retrieval results with 4 baselines on NUS-WIDE
single label subset. (\emph{model})+SG+ML means \emph{(model) }is
integrated with sparse groups and metric learning. \emph{MAP@100}
is evaluated at top 100 similar images. \emph{N@10} = NDCG estimated
at top 10 results. {\scriptsize{}($\uparrow$\%) }denotes improved
percentage.\label{tab:IR_Retrieval_Result_on_NUS-WIDE_single_label_subset.}}

\centering{}{\scriptsize{}}%
\begin{tabular}{|c|c|c|}
\hline 
\multirow{2}{*}{\emph{\footnotesize{}Method}} & \multirow{2}{*}{\emph{\footnotesize{}MAP@100 }{\footnotesize{}($\uparrow$\%)}} & \multirow{2}{*}{\emph{\footnotesize{}N@10 }{\footnotesize{}($\uparrow$\%)}}\tabularnewline
 &  & \tabularnewline
\hline 
\hline 
{\footnotesize{}$k$NN} & {\footnotesize{}0.283} & {\footnotesize{}0.466}\tabularnewline
\hline 
{\footnotesize{}RBM} & {\footnotesize{}0.381\textpm 0.001(+34.6)} & {\footnotesize{}0.565\textpm 0.001(+21.2)}\tabularnewline
\hline 
\hline 
{\footnotesize{}RBM+SG} & {\footnotesize{}0.402\textpm 0.035(+42.1)} & {\footnotesize{}0.584\textpm 0.001(+25.3)}\tabularnewline
\hline 
{\footnotesize{}MV.RBM} & {\footnotesize{}0.455\textpm 0.002 (+60.8)} & {\footnotesize{}0.631\textpm 0.002 (+35.4)}\tabularnewline
\hline 
{\footnotesize{}MV.RBM+SG} & {\footnotesize{}0.483\textpm 0.002 (+70.7)} & {\footnotesize{}0.668\textpm 0.002 (+43.4)}\tabularnewline
\hline 
\textbf{\footnotesize{}MV.RBM+SG+ML} & \textbf{\footnotesize{}0.508\textpm 0.002(+79.5)} & \textbf{\footnotesize{}0.679\textpm 0.001(+45.7)}\tabularnewline
\hline 
\end{tabular}{\scriptsize \par}
\end{table}

Different from the first experiment, we query within testing set for
each testing image\footnote{This way of testing is more realistic since we do not always have
all images for training.}. Table \ref{tab:IR_Retrieval_Result_on_NUS-WIDE_single_label_subset.}
reports the retrieval results of all RBM models. Again, it demonstrates
that (i) representation learning, especially when it comes to fusing
multiple feature types and modalities, is highly important in image
retrieval, (ii) adding structured sparsity can improve the performance,
and (iii) distance metric, when jointly learned with representation,
has significant effect on the retrieval quality. In particular, the
improvement over the $k$-NN when using the proposed method is significant:
MAP score increases by $79.5\%$ and NDCG score by $45.7\%$.

\section{Discussion and Future Works\label{sec:research_proposal}}

\subsection{Drawbacks and Future Plans}

Referring back to previous sections, our principal direction concentrates
on learning a better representation which critically influences the
performance of machine learning methods. However, we have explored
``shallow'' architectures (e.g., generative and discriminative models
with only one hidden layer, Support Vector Machines) which are too
simple to model mixed data. Such shallow architectures are unable
to capture several complex types of multisource data \cite{bengio_ftml09_deep}.
Thus constructing a deep architecture has been obtaining a lot of
attention. One of the typical examples is multilayer neural networks.
Unfortunately, they requires a huge amount of labelled data and often
are trapped at poor local optima.

Hinton~\emph{et~al}. \cite{hinton_neucom06_fastDBN} propose a deep
generative model known as Deep Belief Network (DBN). They also contribute
an unsupervised learning algorithm which helps train DBN quickly.
Later, the generalised model of RBM called Deep Boltzmann Machine
(DBM) is introduced as a deep architecture. Since then, they have
opened the door to a novel research direction called ``deep learning''.
Applications of deep learning then vary in speech recognition \cite{mohamed_etal_2012_acoustic_dbn},
object recognition \cite{hinton_salakhutdinov_sci06_reducing,ciresan_etl_cvpr12_cnn,krizhevsky_nips12_imagenet}
and statistical machine translation \cite{schwenk_hlt12_smt}. In
such fields, deep learning models often achieve state-of-the-art results
or potential outcomes comparable to the supremacy rivals. Additionally,
Deep Auto-Encoder is introduced to enhance the data representations
by fine-tuning in unsupervised fashion. Recently, deep learning has
also been applied to multimodal learning using DBN \cite{ngiam_etal_icml11_multimodal}
and DBM \cite{srivastava_etal_nips12_multimodal_dbm}.

Interestingly, RBM-based models can easily be stacked to form a deep
architecture such as Deep Belief Network (DBN). It also has been proved
that building a deeper model by stacking more hidden layers increases
the log probabilities of the data \cite{hinton_neucom06_fastDBN}.
Besides, the key feature of this deep architecture is greedy layer-wise
training that can be repeated several times to learn a deep hierarchical
model. After initialising weights for all layers by using multiple
RBMs, all parameters of this hierarchical model are fine-tuned by
back-propagation similar to neural networks. The advantages of unsupervised
pretraining with respect to deep architectures are empirically proved
in \cite{erhan_etal_jmlr10_pretrain}. In addition, our proposed models
basing on RBM and MV.RBM demonstrate their capabilities in latent
patient profile modelling in medical data analysis and representation
learning for image retrieval according to Sec.~\ref{sec:apps}. Thanks
to promising perspective, we have grasped motivations and preliminary
guarantees of success to extend our proposed models to deep versions.

Another research direction is to explore temporal information in mixed
data such as sequential frames in video, serial signals in audio and
disease escalation of patients over time. In this field, the mixed-variate
property can be integrated into existing models including conditional
RBM and temporal RBM \cite{taylor_etal_jmlr11_time}.

\subsection{Future Objectives\label{sub:Objectives}}

Our proposed models still focus on learning representation in latent
space for mixed data. The aims are now furthered to learn hierarchical
latent representations for mixed data and integrate temporal information
into deep architecture. The detailed objectives are to:
\begin{itemize}
\item Improve deep architectures to tackle multitype and multimodal data.
The deep here are built for classification, regression and encoding,
decoding purposes;
\item Integrate structured sparsity and distance metric learning into deep
architectures as well as investigate how many levels and which levels
should be enhanced by structured sparsity and distance metric; and
\item Upgrade conditional and temporal RBMs to mixed-variate versions and
build deep architectures basing on these mixed-variate models.
\end{itemize}

\subsection{Significance and benefits}

The significance of our further research can be recognised in two
fields: machine learning and life supports. In machine learning, our
main contributions are: (i) Implementing versatile RBM-based machineries
to handle mixed data; (ii) Novel extension and application of a powerful
data mining tool, Mixed-Variate RBM, to deep architectures and (iii)
Constructing robust hierarchical latent representations with sparsity
structures and distance metric learning. In life supports, the significance
of our work lies in: (i) Building a robust framework that is able
to support healthcare centres and clinicians delivering outcomes that
can integrate with their operations to enhance clinical efficiencies.
Using such systems, the management and supervision on diabetes patients
in particular as well as other kinds of diseases patients in general
would have the potential to improve; (ii) Providing a powerful representation
learning machinery for real applications including face recognition,
object classification and image retrieval.

\subsection{Methodology}

In order to achieve the objectives outlined in Sec.~\ref{sub:Objectives},
we are planning to implement the following methods:
\begin{itemize}
\item \textbf{Mixed-variate deep architecture.} In this architecture, the
lowest layer employs a Mixed-Variate RBM (see Sec.~\ref{sub:MV.RBM})
to model mixed-data with binary units in hidden layer. Next, standard
binary RBMs are stacked to build higher layers. The lowest layer and
higher layers form a block which can be pretrained layer-by-layer
to find good regions in parameter space \cite{hinton_salakhutdinov_sci06_reducing}.
After pretraining stage, there are two approaches for the late training
phase depending on the purpose the model serves. For classification
and regression problems, we add another layer on the recent highest
layer of the deep model. The additional layer contains the output
units such as softmax, logistic units for classification and real
units for regression. For data reconstruction, the recent deep model
is unrolled in order to create an autoencoder. This autoencoder is
in palindromic form with the last layer representing the data. Finally,
following this late training phase, we fine-tune the last model using
backpropagation, minimising error objective function or maximising
log-likelihood function, to figure out optimal set of parameters \cite{hinton_salakhutdinov_sci06_reducing}.
Note that, we must resort to specific objective function for each
type of output units such as Poisson log-likelihood function for count
data, cross-entropy error for binary value and Gaussian log-likelihood
function, the same as mean square error, for real one.
\item \textbf{Deep learning with structured sparsity and distance metric.}
Hierarchical representations can be learned by deep architectures.
These representations are retrieved using the output of units at level-based
layer. Machine learning methods then can be applied to the representation
at the last layer. Obviously, the better the last representation is,
the more superior results we can obtain. According to Sec.~\ref{sub:IR_IntgFramework},
we have successfully integrated structured sparsity and distance metric
learning into the latent representation of MV.RBM. The sparsity and
distance metric are enhanced during the training phase. Similarly,
we can deploy this way to the pretraining and fine-tuning stages of
deep architectures. However, we need to investigate which layers we
should integrate these things into because the integration can be
applied to either any single layer or their combinations.
\item \textbf{Mixed-variate conditional and temporal RBMs.} The standard
RBM-based frameworks only model static frames of data. They have not
taken temporal information into account. Considering temporal as well
as time series factor, the data should be sorted by time order. The
data, then, is splited into timing windows (time slices). We are now
able to model temporal dependencies by treating previous time slices
as additional visible variables. A model called Temporal RBM (TRBM)
is proposed in order to add directed connections from the past configurations
of visible and hidden units \cite{sutskever_etal_aistats07_learning}.
There are three kinds of connections in full TRBM: connections among
the hidden variables, connections among the visible variables and
connections among the visible and hidden variables. Full TRBM can
even have the fourth kind of connections from the past hidden configurations
to the current visible units. In fact, the model is already powerful
enough without the fourth connections. To simplify the full TBRM,
Taylor\emph{~et~al.} \cite{taylor_etal_nips06} ignore previous
hidden configurations and connections among hidden units. They only
add two types of directed connections: autoregressive connections
from the last $N$ states of visible units to the current visible
states, and connections from the last $M$ states of visible units
to the current hidden states. These connections transform RBM into
Conditional RBM (CRBM). Because all states of visible units in CRBM
and TRBM act as visible ones in the standard RBM, we can use each
state of visible units to model mixed data like the way MV.RBM has
performed. Once CRBM and TRBM successfully model mixed data, multiple
layers of them can be stacked to build deep architectures.\end{itemize}

\section{Conclusion\label{sec:conclusion}}

To summarise, we have provided an overview of mixed data analysis
in statistics and machine learning. We have reviewed related literatures
on three main approaches: applying techniques on original features;
non-probabilistically transforming data to higher-level representations
and modelling mixed data using latent variable frameworks. Our main
approaches are to utilise Restricted Boltzmann Machine (RBM) and its
variants, bipartite undirected graphical models with visible layer
observing the data and hidden layer representing the latent representation.
We have introduced firstly, our extensions of parameter sharing and
balancing and secondly, integrated structured sparsity and distance
metric learning into RBM and Mixed-Variate RBM (MV.RBM). We then applied
our proposed models in various applications including \emph{latent
patient profile} modelling in medical data analysis and representation
learning for image retrieval. The experimental results demonstrated
the models performed better than baseline methods in medical data
and outperformed state-of-the-art rivals in image dataset. In future
works, we aim to learn hierarchical latent representations for mixed
data and integrate temporal information into deep architecture. The
significance of our further research would be recognised in two fields:
machine learning and life supports.

\bibliographystyle{plain}
\bibliography{condidature_confirmation}

\end{document}